\pdfoutput=1

\documentclass[11pt]{article}

\usepackage[final]{acl}

\usepackage{times}
\usepackage{latexsym}

\usepackage[T1]{fontenc}

\usepackage[utf8]{inputenc}

\usepackage{microtype}

\usepackage{inconsolata}

\usepackage{graphicx}

\usepackage{amsmath,amssymb}
\usepackage{algorithm}
\usepackage{algpseudocode} 
\definecolor{xbjblue}{RGB}{82,143,173}
\definecolor{xbjred}{RGB}{231,98,84}
\definecolor{grey}{gray}{0} 
\definecolor{xbjgreen}{rgb}{0.0, 0.42, 0.24}
\usepackage{xspace}
\usepackage{subfiles}
\usepackage{booktabs}
\usepackage{multirow}
\usepackage{tabularx}
\usepackage[table]{xcolor}
\usepackage{makecell}   
\usepackage{pifont} 
\usepackage{threeparttable} 
\usepackage{siunitx}    
\usepackage{titlesec}
\usepackage[most]{tcolorbox}
\usepackage{CJKutf8}
\usepackage[dvipsnames]{xcolor}
\usepackage[most]{tcolorbox}

\titlespacing*{\paragraph}{0pt}{4pt}{4pt}
\newcommand{\affpoo}{\textit{Affinity Pooling}\xspace}
\newcommand{\cmark}{\ding{51}} 
\newcommand{\nmark}{\textcolor{gray}{-}} 
\newcolumntype{Y}{>{\centering\arraybackslash}X}
\newcommand{\eos}{%
  \begingroup
  \setlength{\fboxsep}{1pt}
  \color{xbjblue!70!black}%
  \fcolorbox{xbjblue!70!black}{xbjblue!10}{\texttt{<|endoftext|>}}%
  \endgroup
}

\newcommand{\looping}{%
  \begingroup
  \setlength{\fboxsep}{1pt}%
  \fcolorbox{RedOrange!85!black}{Orange!12}{%
    \texttt{\bfseries\color{RedOrange!85!black}\texttt{Looping}}%
  }%
  \endgroup
}

\title{Do We Need Distinct Representations for Every Speech Token?\\Unveiling and Exploiting Redundancy in Large Speech Language Models}

\author{Bajian Xiang\thanks{Corresponding authors.} \quad Tingwei Guo \quad Xuan Chen\footnotemark[1] \quad Yang Han \\
Beike Inc., Beijing, China \\
\texttt{\{xiangbajian001,guotingwei002,chenxuan046,hanyang030\}@ke.com}
\\
\url{https://xchen-zero.github.io/speech-token-redundancy/}
}

\begin{document}
\maketitle

\begin{abstract}

Large Speech Language Models (LSLMs) typically operate at high token rates (tokens/s) to ensure acoustic fidelity, yet this results in sequence lengths that far exceed the underlying semantic content, incurring prohibitive inference costs. In this paper, we empirically revisit the necessity of such granular token-level processing. Through layer-wise oracle interventions, we unveil a structured redundancy hierarchy: while shallow layers encode essential acoustic details, deep layers exhibit extreme redundancy, allowing for aggressive compression. Motivated by these findings, we introduce \affpoo, a training-free, similarity-based token merging mechanism. By strategically applying this method at both input and deep layers, we effectively compress speech representations without compromising semantic information. Extensive evaluations across three tasks demonstrate that our approach reduces prefilling FLOPs by 27.48\% while maintaining competitive accuracy. Practical deployment further confirms significant efficiency gains, yielding up to $\sim$1.7$\times$ memory savings and $\sim$1.1$\times$ faster time-to-first-token on long utterances. Our results challenge the necessity of fully distinct token representations, providing new perspectives on LSLM efficiency.

\end{abstract}

\section{Introduction}

Large Speech Language Models (LSLMs) typically process audio at high token rates to ensure acoustic fidelity \cite{cui2025recentadvancesspeechlanguage-survey1, bu2024roadmapsuperhumanspeechunderstanding-survey2}. 
However, speech naturally exhibits highly non-uniform redundancy over time, resulting in a token sequence that grows far faster than the underlying semantic content \cite{wang2025codecslimetemporalredundancycompression-tokenizer1, zheng2025saylessvariableframeratespeech-tokenizer2}.
This forces the language backbone to process many redundant tokens, incurring substantial and often unnecessary computation.

Similar redundancy has also been observed in Vision-Language Models (VLMs), motivating a line of work on token compression to reduce sequence length while preserving task-relevant semantics \cite{wen2025tokenpruningmultimodallarge, shao2025tokenstalkmuchsurvey}.  
In contrast, compression for LSLMs remains relatively underexplored: existing speech-centric methods largely borrow VLM techniques by operating on spectrograms \cite{behera2024-FastAST-acceleratingaudiospectrogram-AudioPatch1, Lee_2025-AudioPatch2} or apply compression in specialized architectures \cite{li2023acceleratingtransducersadjacenttoken-atome}. More importantly, it remains unclear how redundancy is distributed across layers in LSLMs, hindering principled choices of where and how aggressively to apply compression.

To bridge this gap, we investigate internal redundancy of LSLMs through layer-wise oracle interventions. By dropping or merging tokens based on supervised linguistic boundaries, we unveil a clear hierarchy: shallow layers encode fine-grained details while \textbf{deep layers exhibit extreme redundancy, allowing significant token reduction with negligible performance degradation}. We further analyze the layer-wise dynamics of speech token cosine similarity to explain this behavior.

Building on these findings, we introduce \affpoo, an unsupervised similarity-based compression algorithm. We first validate it as an intervention probe, demonstrating that \textbf{intrinsic similarity captures essential information more effectively than supervised alignment}. We then formalize \affpoo and its variant, \textit{Dual Affinity Pooling} (DAP), as training-free mechanisms applied during inference. Extensive evaluation across three semantic speech tasks confirms that DAP reduces FLOPs by 27.48\% while preserving or improving accuracy. Practical measurements further show consistent deployment gains of up to $\sim$1.7$\times$ memory saving and $\sim$1.1$\times$ faster time-to-first-token (TTFT) on long utterances.

Our contributions are three-fold:
(1) We are the first to anatomize layer-wise redundancy in LSLMs via controlled interventions, offering an interpretable view of their inner workings;
(2) We propose \affpoo, a training-free, similarity-driven token compression algorithm whose design is explicitly grounded in the above analysis;
(3) We validate the approach across multiple models and semantic speech tasks, and conduct targeted sensitivity analyses of key hyperparameters.

\section{Related Work}

\subsection{Large Speech Language Models}

LSLMs have emerged as a prominent paradigm for processing spoken inputs, typically consisting of a speech encoder, an alignment module, and a Large Language Model (LLM) backbone. The speech encoder converts raw audio into token sequences, which vary in design philosophy: continuous encoders such as Qwen2-Audio \cite{qwen2audio} directly extract acoustic features, discrete encoders such as GLM-4-Voice \cite{glm4voice} and Baichuan-Audio \cite{baichuanaudio} quantize speech into symbolic tokens, and hybrid approaches like Kimi-Audio \cite{kimiaudio} combine both representations to leverage their complementary strengths.

Regardless of the architectural choice, these models often operate at high tokenization rates, typically ranging from 12.5 to 25 tokens per second of audio \cite{ji2024wavchatsurveyspokendialogue-survey3}, producing sequences substantially longer than their textual counterparts, inherently suggesting a high degree of redundancy.

\subsection{Token Compression in Multimodal LLMs}

To mitigate the computational overhead of high-resolution inputs in VLMs, the vision-language community has developed diverse compression strategies, ranging from attention-based pruning \cite{yang2024-visionzip-vision, shao2025-holitom-video} to similarity-driven merging \cite{bolya2023-tome-vision, tao2025-dycoke-video}. Recent advancements have transcended heuristic methods, prioritizing interpretability to rigorously identify which representations are suitable to pruning or merging \cite{fu2025-framefusion-vision}.

Token compression for LSLMs remains underexplored. 
Recent attempts like SpeechPrune \cite{lin2025speechprunecontextawaretokenpruning-speechprune} and TimeAudio \cite{wang2025listeningframesbridgingtemporal-timeaudio} utilize attention-guided pruning or learnable aggregation modules before the LLM input. These initial explorations, while promising, operate primarily through empirical design without grounding in speech-specific representational analysis.

\begin{figure*}[h]
    \centering
    \includegraphics[width=1\linewidth]{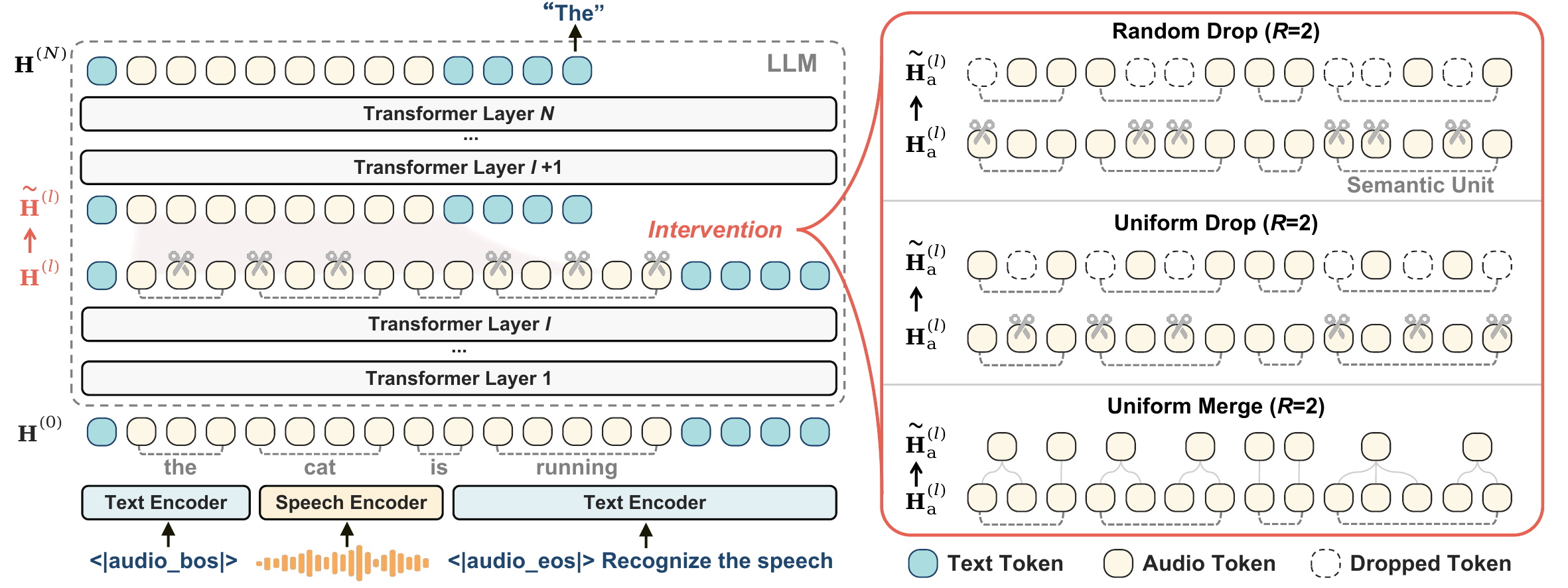}
    \captionsetup{skip=3pt}
    \caption{\textbf{Framework of oracle intervention experiments.} We align audio tokens to semantic units and apply compression operators \textbf{to a single layer at a time} to investigate redundancy.}
    \label{fig:3_1_intervention}
\end{figure*}

\section{Anatomy of Redundancy Through Oracle Interventions}
\label{Chap3}

This section investigates the redundancy of audio token representations within LSLMs. We use word-level timestamps as an Oracle to align the audio token stream with its corresponding linguistic units, and then compress the token stream within specific semantic windows through dropping or merging. By assessing the recoverability of the original content from these reduced sequences via an ASR task, we empirically quantify the structural redundancy inherent in the audio representation.


\subsection{Methodology}
\label{sec3_1_expdetail}
\paragraph{Intervention Framework.}
We formally define the intervention process within the latent space of the LSLM, as illustrated in Figure~\ref{fig:3_1_intervention}. Let $\mathbf{H}^{(l)} = [\mathbf{H}_a^{(l)}; \mathbf{H}_t^{(l)}] \in \mathbb{R}^{(T_a + T_t) \times d}$ denote the concatenated hidden states at layer $l$, where $T_a$ and $T_t$ represent the sequence lengths of audio and text, respectively, and $d$ is the hidden dimension. Our objective is to apply a compression operator $\Phi$ solely to the audio component, yielding a reduced representation $\tilde{\mathbf{H}}_a^{(l)} = \Phi(\mathbf{H}_a^{(l)})$, while leaving the text component $\mathbf{H}_t^{(l)}$ intact. The subsequent layer $l{+}1$ then processes the modified sequence $[\tilde{\mathbf{H}}_a^{(l)}; \mathbf{H}_t^{(l)}]$.

\paragraph{Compression Strategies.}
To quantify redundancy, we partition the audio sequence into semantic units aligned with word-level timestamps. We then compress each unit to a fixed budget of $R$ tokens via three operators: (1) \textit{Random Drop}, which stochastically samples $R$ tokens; (2) \textit{Uniform Drop}, which samples deterministically at regular strides to preserve temporal structure; and (3) \textit{Uniform Merge}, which divides the unit into $R$ equal-sized bins and performs mean-pooling within each bin.

\paragraph{Experimental Setup.}
We utilize Qwen2-Audio (32 layers, 25 tokens/s) and Kimi-Audio (28 layers, 12.5 tokens/s) on the Librispeech-test-clean test set. We evaluate the semantic recoverability of the compressed representations via Word Error Rate (WER) on an ASR task. For generation, we employ greedy decoding with a maximum token limit of 256. Word alignments are obtained offline via the Montreal Forced Aligner (MFA). To trace the layer-wise evolution of redundancy, we apply interventions \textbf{to single layers individually} at intervals of five. The retention budget $R$ is scaled according to frame rates: $R \in \{2, 4, 8, 16\}$ for Qwen2-Audio and $R \in \{1, 2, 4, 8\}$ for Kimi-Audio.

\begin{figure*}[t]
    \centering
    \includegraphics[width=1\linewidth]{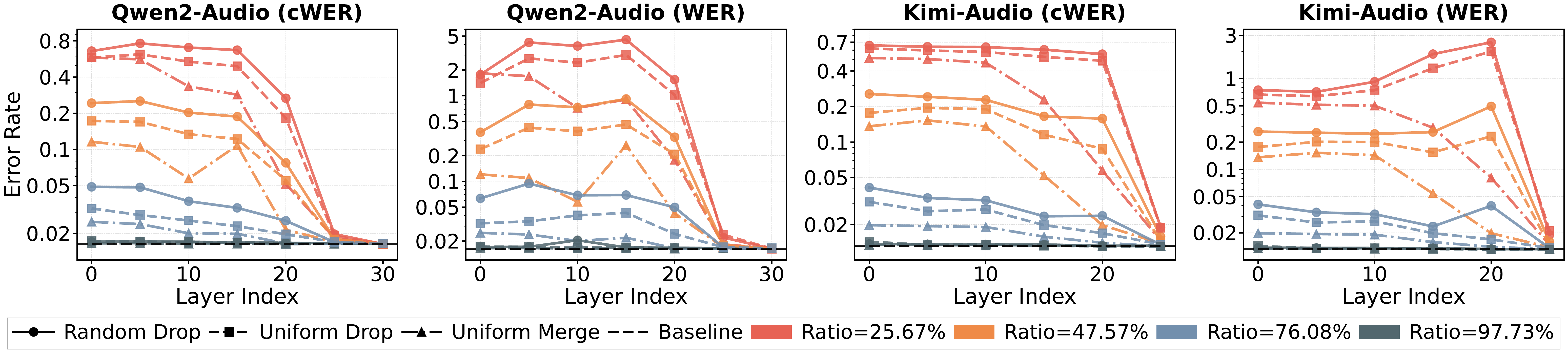}
    \caption{
        \textbf{Layer-wise oracle interventions on Qwen2-Audio and Kimi-Audio}.
        For each model, we report clamped WER (cWER) and standard WER plotted on \textbf{log-scale}. Colors represent different audio token retention rates.
    }
    \label{fig:chap3_layerwise_main}
\end{figure*}

\subsection{Layer-wise Redundancy Evolution}
\label{sec:layerwise_analysis}

Figure \ref{fig:chap3_layerwise_main} illustrates the intervention results, where curves of different colors correspond to varying retention budgets $R$ (converted here into audio token retention ratios).
To isolate semantic loss from degenerate decoding behaviors, we report standard WER alongside clamped WER (cWER).
Formally, for a dataset with samples indexed by $i$, we define $\text{cWER} = \sum_i \min(E_i, N_i) \,/\, \sum_i N_i$, where $E_i$ and $N_i$ are the edit distance and reference length for the $i$-th sample.
Full numerical results are provided in Appendix \ref{app:oracle_details}.
We distill our observations into three primary findings:

\noindent(1) \textbf{Progressive Growth of Redundancy.} As evidenced by the cWER profiles, the model's sensitivity to token removal decreases monotonically with depth. While shallow layers require high retention budgets to preserve acoustic fidelity, deep layers ($l \ge 25$) exhibit extreme redundancy; performance converges to the baseline even when retaining as few as 25.67\% of the original tokens. This observation suggests that deeper layers may harbor a significantly higher degree of redundancy.

\noindent(2) \textbf{Acoustic-to-Semantic Transition.} A large gap between WER and cWER appears in the middle layers ($l \in [5, 15]$ for Qwen2-Audio; $l \in [15, 20]$ for Kimi-Audio). This gap is primarily caused by repetition loops, as shown in the decoding examples in Table~\ref{tab:loop_of_qwen} and Table~\ref{tab:loop_of_kimi} (Appendix~\ref{app:oracle_samples}). Besides loops, we also observe unstable behaviors like cross-lingual hallucinations and semantic drift. For instance, the model paraphrases "ghost" to "spirit" or "vesture" to "veil." This suggests that the middle layers are in a critical transitional state. \textbf{The model has started to abstract high-level semantics from acoustic features} but has not yet fully aligned them with exact lexical tokens.

\noindent(3) \textbf{Structural Nature of Redundancy.} The hierarchy of intervention strategies highlights that speech redundancy is not random. Across all layers, \textit{Uniform Drop} consistently outperforms \textit{Random Drop}, suggesting that redundancy possesses a temporal structure that benefits from regular sampling. Furthermore, \textit{Uniform Merge} yields the lowest error rates consistently across both models and varying retention budgets. This indicates that tokens deemed "redundant" likely still contain distributed information; consequently, aggregating these features via averaging appears to preserve semantic cues more effectively than simple excision.


\begin{figure}
    \centering
    \includegraphics[width=\linewidth]{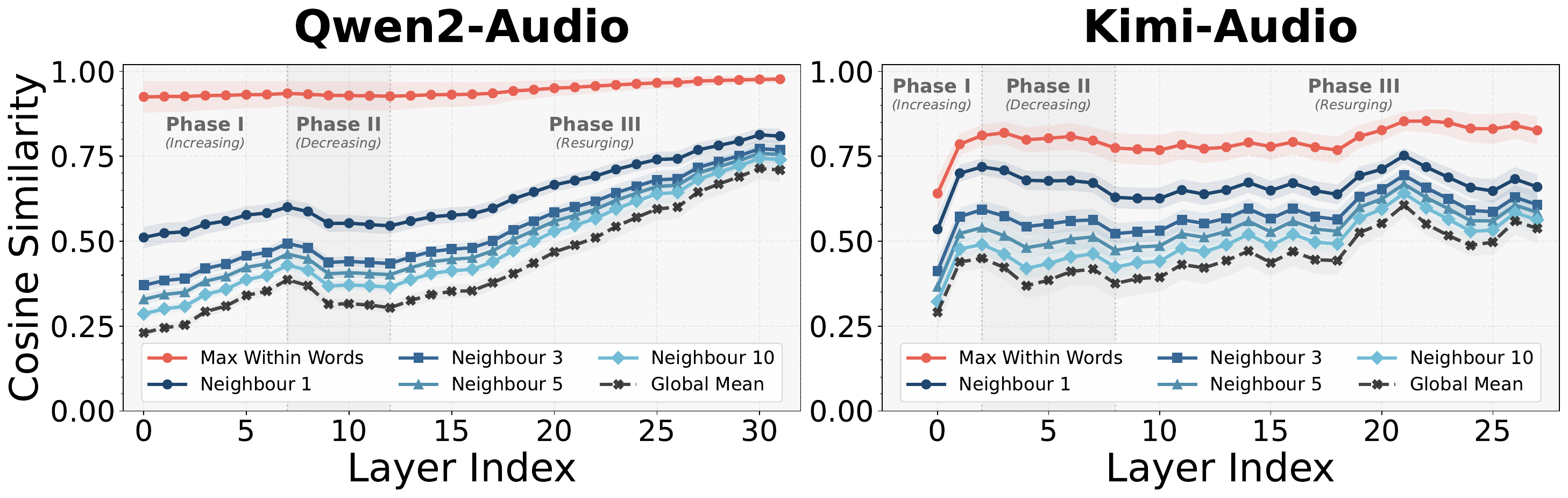}
    \captionsetup{skip=3pt}
    \caption{Layer-wise cosine similarity dynamics for Qwen2-Audio and Kimi-Audio.}
    \label{fig:chap3_cos_dynamics}
\end{figure}

\subsection{Deep Dive: Feature Dynamics}
\label{sec:feature_dynamics}

To understand the intrinsic mechanisms driving the observed redundancy, we analyze the layer-wise evolution of audio representations on the Librispeech-test-clean set. As shown in Figure \ref{fig:chap3_cos_dynamics}, we track three cosine similarity metrics: (1) Neighbor Similarity, the average similarity between a token and its $k$-nearest neighbors ($k{\in}\{1, 3, 5, 10\}$); (2) \textit{Global Mean}, the average similarity across the entire sequence; (3) \textit{Max Within Words}, the maximum adjacent similarity within word boundaries.

The unsupervised metrics display a clear rise–fall–rise trajectory. Phase II indicates that local relationships between nearby tokens become less consistent, suggesting a period of representation reorganization. This aligns with the high intervention sensitivity noted in Section~\ref{sec:layerwise_analysis}. We interpret this instability as a symptom of the critical transition from dense acoustic features to the highly redundant semantic abstractions emerging in deeper layers. Notably, Kimi-Audio exhibits a metric drop in Phase III, likely due to its unique design reusing layer 21 for acoustic decoding.

\textit{Max Within Words} shows an overall upward trend and typically peaks in deep layers; Qwen2-Audio also displays a notably high \textit{Global Mean} near the top. These metrics indicate that deep layers map tokens within the same linguistic unit to highly similar vectors, creating significant representational redundancy, which explains why aggressive merging becomes effective at depth.

\section{Similarity–Driven Interventions}
While the previous section identified \textbf{when} to compress by pinpointing redundant layers, this section investigates \textbf{where} to compress by exploring the specific token structures governing this redundancy. We introduce \affpoo, a method that aggregates tokens based on feature cosine similarity. We first detail the algorithm, evaluate its efficiency against the oracle baseline, and then interpret the semantic granularity of the merged tokens.


\subsection{\affpoo}
\label{sec:affinity_pooling}

Building on the findings in Section \ref{Chap3}, we propose \affpoo (Algorithm \ref{alg:affinity_pooling}), where the term \textit{affinity} captures the semantic closeness between token representations as measured by cosine similarity.

Our design diverges from existing paradigms in the following aspects: Unlike vision-centric global matching \cite{bolya2023-tome-vision}, our design strictly adheres to the intrinsic \textbf{temporal locality} of speech. Furthermore, distinct from heuristic adaptations in prior speech works \cite{li2023acceleratingtransducersadjacenttoken-atome}, our use of latent similarity is explicitly grounded in the structural redundancy revealed in Section \ref{sec:feature_dynamics}.
We also address the limitations of standard adjacent merging by introducing a \textbf{lookback window} $\omega$.
While strict adjacency ($\omega=1$) is susceptible to high-frequency acoustic jitter, our windowed approach ($\omega > 1$) bridges local fluctuations, preserving semantic continuity without compromising the similarity threshold $\tau$. 

Operationally, a token $h_t$ is merged into the active group if its cosine similarity with any of the most recent $\omega$ tokens exceeds $\tau$; otherwise, the current group is aggregated via mean-pooling.


\begin{algorithm}[t]
\small
\caption{Pseudocode for \affpoo}
\label{alg:affinity_pooling}
\begin{algorithmic}[1]
\State \textbf{Input:} Audio sequence $\mathbf{H}_a = [h_1, \dots, h_{T_a}] \in \mathbb{R}^{T_a \times d}$, lookback window size $\omega$, similarity threshold $\tau$
\State \textbf{Output:} Merged sequence $\tilde{\mathbf{H}}_a$

\State $\tilde{\mathbf{H}}_a \gets \emptyset$

\State $\mathcal{G}_{curr} \gets [h_1]$ 
\Comment{initialize current group}
\For{$t = 2$ to $T_a$}
    \State $k \gets \min(|\mathcal{G}_{curr}|,\ \omega)$ 
    \State $\mathbf{K} \gets$ last $k$ tokens in $\mathcal{G}_{curr}$ 
    \State $s_{\max} \gets \max_{k_i \in \mathbf{K}} \cos(h_t, k_i)$ 
    \If{$s_{\max} \ge \tau$}
        \State Append $h_t$ to $\mathcal{G}_{curr}$
    \Else
        \State $\bar{h} \gets \frac{1}{|\mathcal{G}_{curr}|} \sum_{h \in \mathcal{G}_{curr}} h$ 
        \Comment{mean-pool group}
        \State Append $\bar{h}$ to $\tilde{\mathbf{H}}_a$
        \State $\mathcal{G}_{curr} \gets [h_t]$
    \EndIf
\EndFor

\State $\bar{h} \gets \frac{1}{|\mathcal{G}_{curr}|} \sum_{h \in \mathcal{G}_{curr}} h$
\State Append $\bar{h}$ to $\tilde{\mathbf{H}}_a$
\State \Return $\tilde{\mathbf{H}}_a$
\end{algorithmic}
\end{algorithm}


\begin{figure}[h]
    \centering
    \includegraphics[width=\linewidth]{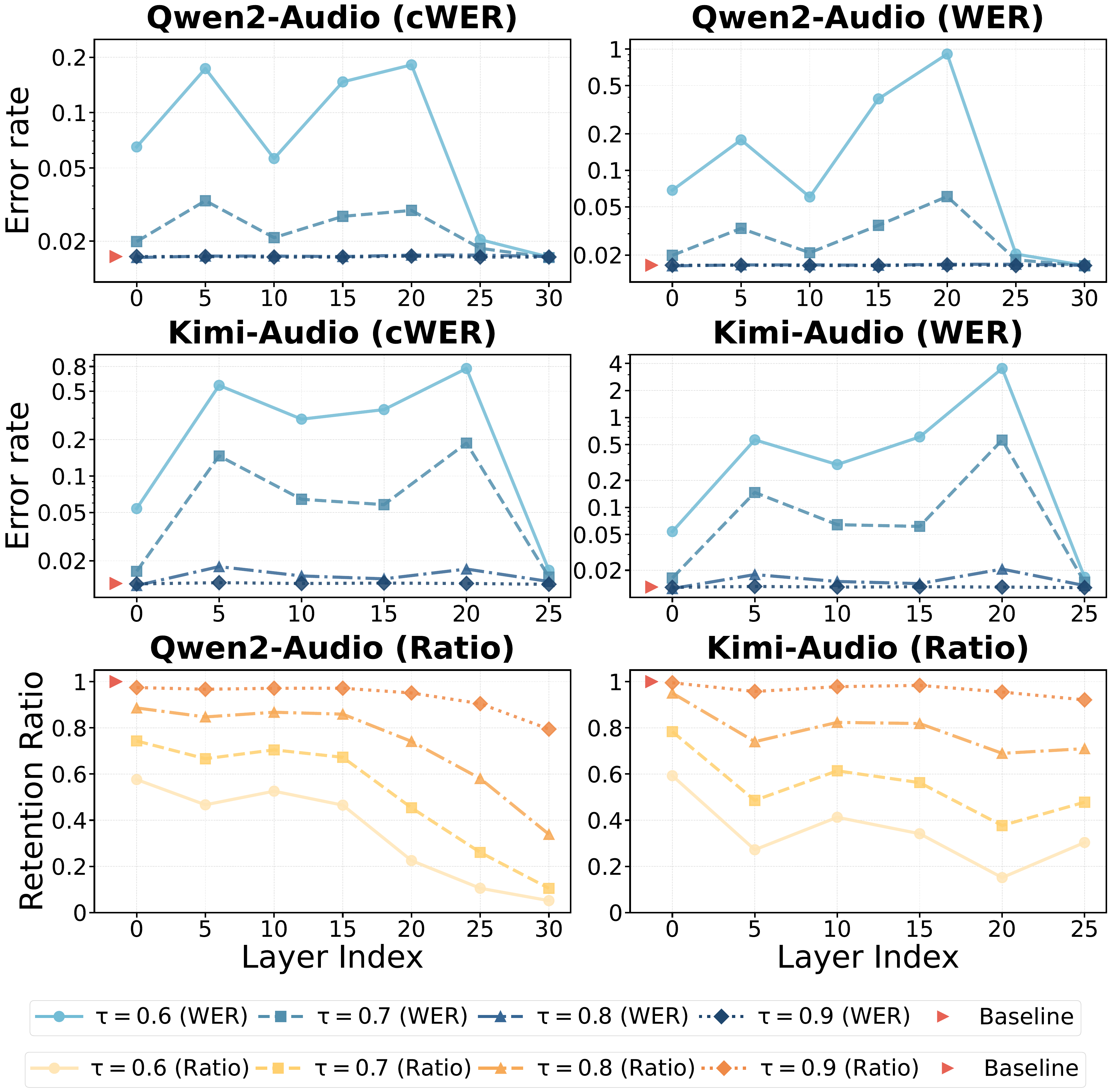}
    \captionsetup{skip=3pt}
    \caption{Layer-wise dynamics of \affpoo on Qwen2-Audio and Kimi-Audio. We report WER, cWER on log-scale, and retention ratios with $\omega=3$ across varying thresholds $\tau \in \{0.6, 0.7, 0.8, 0.9\}$.}
    \label{fig:chap4_2_wer_ratio_affpoo}
\end{figure}

\subsection{Layer-wise Dynamics}
\label{sec4_2_layerwise_dynamics}

We evaluate \affpoo on Qwen2-Audio and Kimi-Audio separately to individual layers under the setup in Section \ref{sec3_1_expdetail}. We fix $\omega=3$ here as an empirical default, focusing on probing layer-wise redundancy, while deferring sensitivity analysis to Section \ref{sec:param_analysis}. Figure \ref{fig:chap4_2_wer_ratio_affpoo} (data in Appendix \ref{app:similarity_results}) illustrates the results, revealing three key phenomena:

\noindent(1) \textbf{Non-monotonic feature stability.}
As illustrated in Figure \ref{fig:chap4_2_wer_ratio_affpoo}, \affpoo exhibits a bimodal error profile across both WER and cWER metrics. While representations at the input ($l=0$) and deep layers ($l \ge 25$) remain robust, intermediate depths degrade significantly, showing distinct error spikes around $l=5$ and $l=20$ under aggressive thresholds ($\tau < 0.8$). This sensitivity mirrors the trend observed in the Oracle baseline as discussed in Section \ref{sec:layerwise_analysis}, confirming that intermediate layers undergo critical feature reorganization and are sensitive and best left uncompressed.

\noindent(2) \textbf{Substantial compression at deep layers.}
Retention ratios consistently decrease as depth increases (Figure \ref{fig:chap4_2_wer_ratio_affpoo}, bottom row). Notably, applying \affpoo to Qwen2-Audio at $l=30$ ($\tau=0.6$) compresses the sequence to 5.18\% of its original length while achieving a WER of 1.64\%, slightly outperforming the uncompressed baseline of 1.65\%. This result indicates that \textbf{deep layers possess significantly higher compressibility than previously observed} in Section \ref{sec:layerwise_analysis}, suggesting that \affpoo effectively uncovers the latent redundancy within these representations.

\noindent(3) \textbf{Superiority over supervised alignment.}
Our unsupervised approach outperforms the supervised Oracle at the input level. At $l=0$ ($\tau=0.7$), we achieve a lower WER of 1.99\% with 74.32\% retention, compared to the Oracle's 2.50\% WER at 76.08\% retention. This suggests that intrinsic similarity captures essential information more effectively than the rigid linguistic boundaries.

\begin{figure*}[!htbp]
    \centering
    \includegraphics[width=1\linewidth]{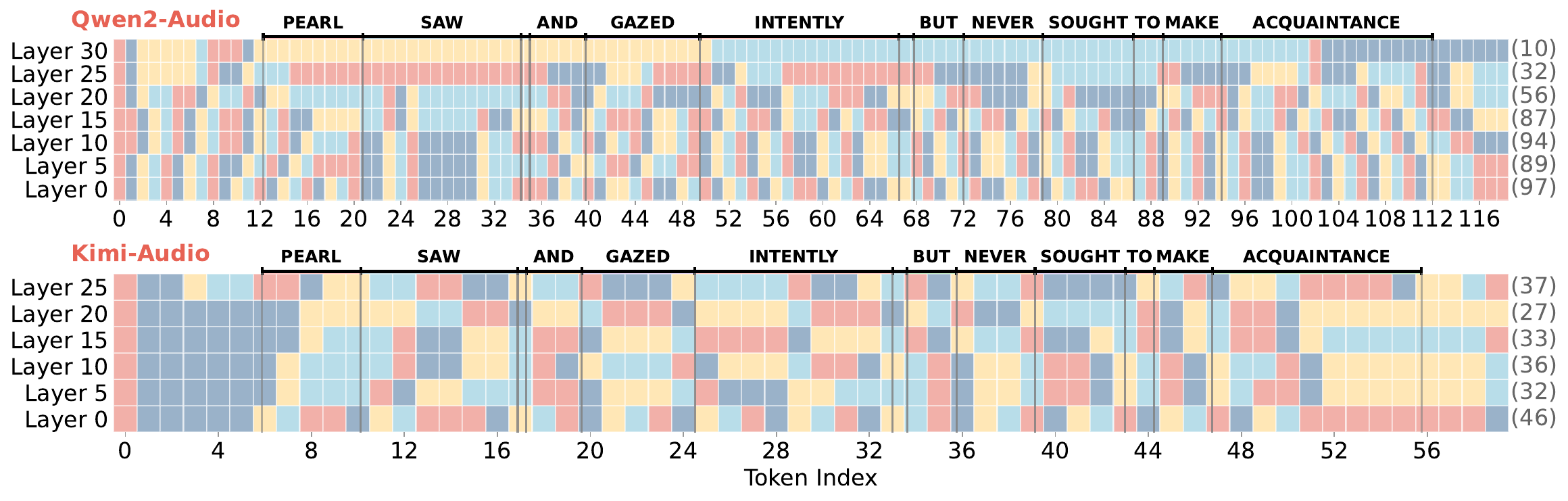}
    \captionsetup{skip=3pt}
    \caption{\textbf{Visualization of \affpoo (\(\tau{=}0.7, \omega=3\)) on Qwen2-Audio (top) and Kimi-Audio (bottom)}. Colors denote merged token groups, and vertical lines mark word boundaries. The right axis indicates the total number of tokens after compression. \textbf{Both models maintain a WER of 0 across all tested layers}.}
    \label{fig:example_of_token_groups}
\end{figure*}

\subsection{Semantic Granularity of Merged Tokens}
\label{sec:qualitative_analysis}

\textit{Why does a simple cosine similarity-based merging strategy achieve such high compression rates while preserving model performance?} To answer this, we analyze the emergent token groups formed during the merging process. Here, we present a representative sample to intuitively illustrate this behavior, while more examples are detailed in Appendix \ref{app:more_visualization}.

Figure \ref{fig:example_of_token_groups} illustrates the layer-wise evolution of token aggregation. We observe a transition from fragmented, acoustic-level groupings in shallow layers ($l \le 5$) to broad semantic abstractions in deep layers ($l \ge 25$). For instance, Kimi-Audio forms continuous token blocks, whereas Qwen2-Audio consolidates larger multi-word blocks, often merging sequences of 4--5 words or more into a single group. Crucially, this structural aggregation preserves fidelity: \textbf{both models achieve a WER of 0 on this utterance across all layers visualized in the figure}. These patterns align with recent findings that LLM embeddings operate far below their theoretical information capacity—where a single vector could encode over 1,500 tokens \cite{kuratov2025cramming1568tokenssingle}. Our method leverages this by merging these adjacent similar tokens, effectively densifying information without exceeding the vector's capacity.

\section{\affpoo for Efficient LSLMs}
In this section, we apply \affpoo as a training-free compression mechanism for LSLMs. We first evaluate the method across multiple downstream tasks, verifying that it maintains high performance despite significant token reduction. To demonstrate practical utility, we report improvements in inference speed and memory consumption on standard hardware. We then benchmark our approach against fixed-budget baselines to highlight its superiority over naive compression methods. Finally, we analyze the impact of key hyperparameters to provide optimal configuration guidelines.

\begin{table*}[t]
\centering
\small

\setlength{\tabcolsep}{2.2pt} 
\renewcommand{\arraystretch}{1.18}

\begin{threeparttable}
\caption{\textbf{Main results and ablation studies on Qwen2-Audio.} FRR: Final Retention Ratio. Bold indicates improvement over Vanilla, while underlining denotes the best performance within each setting.}
\label{tab:qwen_main}

\begin{tabularx}{\textwidth}{l Y Y | c c c | Y Y Y c | c c c c |c c Y}
\toprule
\multirow{2}{*}{\textbf{Method}} &
\multicolumn{2}{c|}{\textbf{Scope}} & 
\multicolumn{3}{c|}{Efficiency ($\downarrow$)} &
\multicolumn{4}{c|}{ASR (WER $\downarrow$)} &
\multicolumn{4}{c|}{QA (Acc $\uparrow$)} &
\multicolumn{3}{c}{ST (BLEU $\uparrow$)} \\
\cmidrule(lr){2-3}\cmidrule(lr){4-6}\cmidrule(lr){7-10}\cmidrule(lr){11-14}\cmidrule(lr){15-17}
& $l_{\mathrm{in}}$ & $l_{\mathrm{deep}}$ 
& \makecell{FRR} & \makecell{Pre.\\GFLOPs} & \makecell{FLOPs\\Ratio}
& KES & LSC & LSO & \textit{\textbf{Avg.}}
& OBQA & SDQA & TrQA & \textit{\textbf{Avg.}}
& en2zh & zh2en & \textit{\textbf{Avg.}} \\
\midrule

Vanilla & \nmark & \nmark
& 100.0 & 780.94 & 100.0
& 3.28 & 1.65 & 3.88 & 2.94
& 42.64 & 27.31 & 21.29 & 30.41
& 42.80 & 23.02 & 32.91 \\

\midrule

\rowcolor{xbjblue!15}
\multicolumn{17}{l}{\textit{Setting A: Aggressive ($\tau_{\mathrm{in}}{=}0.80,\tau_{\mathrm{deep}}{=}0.70$)}} \\

$\text{AP}_{\text{in}}$ & \cmark & \nmark
& 78.64 & 612.93 & 78.49
& 3.39 & \textbf{1.64} & \textbf{3.80} & \textbf{2.94}
& \textbf{43.74} & 26.29 & 20.90 & 30.31
& 42.35 & 22.69 & 32.52 \\

$\text{AP}_{\text{deep}}$ & \nmark & \cmark
& \underline{14.30} & 718.12 & 91.96
& 3.31 & \textbf{1.65} & \textbf{3.84} & \textbf{2.93}
& \textbf{42.86} & \textbf{27.49} & 21.19 & \textbf{30.51}
& 42.76 & \textbf{23.02} & 32.89 \\

\rowcolor{grey!5}
DAP & \cmark & \cmark
& 14.91 & \underline{566.30} & \underline{72.52}
& 3.44 & \textbf{1.63} & \textbf{3.79} & 2.95
& 42.20 & \textbf{29.66} & 20.61 & \textbf{30.82}
& 42.29 & 22.78 & 32.54 \\

\midrule

\rowcolor{xbjblue!15}
\multicolumn{17}{l}{\textit{Setting B: Conservative ($\tau_{\mathrm{in}}{=}0.90,\tau_{\mathrm{deep}}{=}0.80$)}} \\

$\text{AP}_{\text{in}}$ & \cmark & \nmark
& 93.56 & 730.28 & 93.51
& \textbf{3.26} & \textbf{1.65} & \textbf{3.81} & \textbf{2.91}
& \textbf{43.52} & \textbf{27.67} & \textbf{21.58} & \textbf{30.92}
& \textbf{42.84} & \textbf{23.06} & \textbf{32.95} \\

$\text{AP}_{\text{deep}}$ & \nmark & \cmark
& \underline{33.29} & 731.96 & 93.73
& 3.31 & 1.66 & \textbf{3.83} & \textbf{2.93}
& \textbf{44.62} & 26.94 & 21.09 & \textbf{30.88}
& \textbf{42.80} & 23.01 & \textbf{32.91} \\

\rowcolor{grey!5}
DAP & \cmark & \cmark
& 33.76 & \underline{686.39} & \underline{87.89}
& 3.29 & \textbf{1.64} & \textbf{3.77} & \textbf{2.90}
& \textbf{42.64} & \textbf{27.85} & 20.90 & \textbf{30.46}
& \textbf{42.86} & 22.94 & 32.90 \\

\bottomrule
\end{tabularx}
\end{threeparttable}
\end{table*}

\subsection{Performance on Downstream Tasks}
We assess the efficacy of \affpoo across three diverse speech tasks: Automatic Speech Recognition (ASR), Speech Question Answering (QA), and Speech Translation (ST). Our primary objective is to determine whether the proposed method can reduce computational overhead without compromising task performance.

\paragraph{Experimental Setup.}
We utilize Qwen2-Audio as the primary testbed. We compare the uncompressed baseline (Vanilla) against three variants of our method: \affpoo applied only at the input ($\text{AP}_{\text{in}}$), only at a deep layer ($\text{AP}_{\text{deep}}$), and the combined approach, \textit{Dual Affinity Pooling} (DAP). To investigate the tradeoff between efficiency and accuracy, we introduce two configurations for the cosine similarity threshold $\tau$: a \textit{Conservative} setting ($\tau_{\mathrm{in}}{=}0.9, \tau_{\mathrm{deep}}{=}0.8$) prioritizing performance preservation, and an \textit{Aggressive} setting ($\tau_{\mathrm{in}}{=}0.8, \tau_{\mathrm{deep}}{=}0.7$) aiming for maximum compression. For all other hyperparameters, we fix the layer indices at $l_{\mathrm{in}}=0$ and $l_{\mathrm{deep}}=29$, and the window sizes at $\omega_{\mathrm{in}}=1$ and $\omega_{\mathrm{deep}}=3$. All experiments utilize greedy decoding.

\paragraph{Datasets and Metrics.}
We evaluate on multiple benchmarks:(1) ASR: KeSpeech \cite{Tang2021KeSpeechAO} and LibriSpeech \cite{librispeech} (WER); (2) QA: OpenBookQA, SDQA \cite{OBQA_SDQA_voicebench}, and SpeechTriviaQA \cite{SpeechTriviaQA_from_UltraEval} (Accuracy); (3) ST: CoVost2 \cite{Covost2} en2zh and zh2en (BLEU).
For efficiency, we report prefilling GFLOPs and Final Retention Ratio (FRR), defined as the percentage of audio tokens remaining after all compression stages. Further details on dataset specifications and evaluation protocols are provided in Appendix \ref{app:benchmarking} and \ref{app:evaluation details}, respectively.

\paragraph{Main Results.}
Table \ref{tab:qwen_main} presents the performance and efficiency results. \affpoo achieves substantial computational savings with negligible degradation across all tasks. In the \textit{Aggressive} setting, DAP reduces the FRR to 14.91\% and cuts prefilling GFLOPs by 27.48\%. Despite this drastic reduction, it maintains comparable WER and BLEU scores to the baseline, while even improving QA accuracy. Notably, under the \textit{Conservative} setting, $\text{AP}_{\text{in}}$ outperforms the Vanilla baseline on all three task averages, while $\text{AP}_{\text{deep}}$ and DAP also surpass Vanilla in both ASR and QA tasks. We verify these findings generalize to Kimi-Audio, with full results in Appendix \label{app:kimi_main_exp} (Table \ref{tab:kimi_main}).

We observe that while $\text{AP}_{\text{deep}}$ achieves more aggressive token reduction, $\text{AP}_{\text{in}}$ delivers greater computational savings due to cumulative effects across the entire network. This advantage is threshold-dependent: under \textit{Aggressive} settings, $\text{AP}_{\text{in}}$ reduces FLOPs by 21.51\% compared to 8.04\% for $\text{AP}_{\text{deep}}$; under \textit{Conservative} settings, both yield comparable efficiency gains ($\sim$6\%). Notably, DAP's FRR slightly exceeds $\text{AP}_{\text{deep}}$ alone in the \textit{Aggressive} setting, suggesting that early compression may disrupt some long-range redundancies captured at deeper layers. Nevertheless, DAP offers a favorable balance between efficiency and performance across our evaluation suite.

\begin{table}[t]
\centering
\small
\setlength{\tabcolsep}{2pt}
\renewcommand{\arraystretch}{1.1}

\begin{threeparttable}
\captionsetup{skip=3pt}
\caption{\textbf{Prefilling efficiency on H200.}
Time-to-first-token (TTFT, ms), peak memory $m$, and dynamic increment $\Delta m$ (GB), 
where $\mathrm{Spd.}=\mathrm{TTFT}_{\text{Vanilla}}/\mathrm{TTFT}$ and 
$\mathrm{Sav.}=\Delta m_{\text{Vanilla}}/\Delta m$.}

\label{tab:lat_mem_ablation_compact}

\begin{tabularx}{\columnwidth}{ l | Y Y Y | Y Y Y }
\toprule
\multirow{2}{*}{\textbf{Method}} &
\multicolumn{3}{c|}{\textbf{Time Efficiency}} &
\multicolumn{3}{c}{\textbf{Memory Usage}} \\
\cmidrule(lr){2-4}\cmidrule(l){5-7}

& \makecell[c]{TTFT \\ \scriptsize{(ms)}$\downarrow$} 
& \makecell[c]{Spd. \\ \scriptsize{$\uparrow$}} 
& \makecell[c]{$t_{\mathrm{AP}}$ \\ \scriptsize{(ms)}$\downarrow$}
& \makecell[c]{$m$ \\ \scriptsize{(GB)}$\downarrow$} 
& \makecell[c]{$\Delta m$ \\ \scriptsize{(GB)}$\downarrow$} 
& \makecell[c]{Sav. \\ \scriptsize{$\uparrow$}} \\

\midrule
\rowcolor{xbjblue!15}
\multicolumn{7}{l}{\textit{Duration bucket: 40--60 s}} \\
Vanilla & 132.24 & 1.00$\times$ & - & 33.40 & 1.99 & 1.00$\times$ \\
$\text{AP}_{\text{in}}$   & 117.36 & 1.13$\times$ & 0.43 & 33.10 & 1.68 & 1.18$\times$ \\
$\text{AP}_{\text{deep}}$ & 131.31 & 1.01$\times$ & 8.36 & 32.77 & 1.36 & 1.46$\times$ \\
\rowcolor{grey!5}
\textbf{DAP} & 117.87 & 1.12$\times$ & 7.33 & 32.58 & 1.17 & 1.70$\times$ \\

\midrule
\rowcolor{xbjblue!15}
\multicolumn{7}{l}{\textit{Duration bucket: 20--40 s}} \\
Vanilla & 89.90 & 1.00$\times$ & - & 32.53 & 1.15 & 1.00$\times$ \\
$\text{AP}_{\text{in}}$  & 83.58 & 1.08$\times$ & 0.47 & 32.39 & 1.01 & 1.14$\times$ \\
$\text{AP}_{\text{deep}}$ & 89.70 & 1.00$\times$ & 4.34 & 32.16 & 0.78 & 1.47$\times$ \\
\rowcolor{grey!5}
\textbf{DAP} & 84.08 & 1.07$\times$ & 4.31 & 32.08 & 0.70 & 1.64$\times$ \\

\midrule
\rowcolor{xbjblue!15}
\multicolumn{7}{l}{\textit{Duration bucket: 0--20 s}} \\
Vanilla & 53.98 & 1.00$\times$ & - & 31.79 & 0.41 & 1.00$\times$ \\
$\text{AP}_{\text{in}}$   & 54.01 & 1.00$\times$ & 0.43 & 31.76 & 0.38 & 1.08$\times$ \\
$\text{AP}_{\text{deep}}$ & 55.74 & 0.97$\times$ & 1.66 & 31.66 & 0.28 & 1.46$\times$ \\
\rowcolor{grey!5}
\textbf{DAP} & 55.50 & 0.97$\times$ & 2.01 & 31.65 & 0.27 & 1.52$\times$ \\

\bottomrule
\end{tabularx}

\end{threeparttable}
\end{table}

\subsection{Real-World Efficiency}
\label{sec:real_world_efficiency}
To assess practical deployment viability, we measure inference latency and memory consumption on a single NVIDIA H200 GPU. We utilize the Qwen2-Audio model under the \textit{Aggressive} configuration ($\tau_{\mathrm{in}}{=}0.8, \tau_{\mathrm{deep}}{=}0.7$) to establish an upper bound on potential efficiency gains. We curate a test set partitioned into three duration buckets: $D_1$ (0--20s), $D_2$ (20--40s), and $D_3$ (40--60s), with $n=100$ randomly selected samples per bucket.

Since \affpoo specifically optimizes the prompt processing stage, we focus our evaluation on prefilling metrics. We report the time-to-first-token (TTFT), representing the total wall-clock time from input to the first generated token, and explicitly isolate the computational overhead of our algorithm ($t_{\mathrm{AP}}$). For memory, we track the peak VRAM usage ($m$) and the dynamic memory increment ($\Delta m$).

As shown in Table \ref{tab:lat_mem_ablation_compact}, latency gains become more noticeable for longer utterances. Across all three buckets, $\text{AP}_{\text{in}}$ consistently provides the largest wall-clock TTFT speedup, suggesting that early token reduction is a major contributor to end-to-end latency reduction. For long inputs ($D_3$), all compression scopes yield measurable gains, with DAP reaching up to $\sim1.1\times$ speedup. In contrast to latency, DAP consistently reduces GPU memory across all duration buckets. The gain is most pronounced for long utterances ($D_3$), where the dynamic memory increment drops from 1.99\,GB to 1.17\,GB, corresponding to a $\sim1.7\times$ memory saving.


\begin{table}[t]
\centering
\small
\setlength{\tabcolsep}{2.4pt}
\renewcommand{\arraystretch}{1}

\caption{\textbf{ASR results under different token budgets.} WER ($\downarrow$).
Underlines denote the best performance within each budget, and boldface indicates results better than the Vanilla baseline.
}
\label{tab:asr_budget}

\begin{tabularx}{\columnwidth}{l|*{4}{>{\centering\arraybackslash}X}}
\toprule
\textbf{Method} & \textbf{KES} & \textbf{LSC} & \textbf{LSO} & \textit{\textbf{Avg.}} \\
\midrule

Vanilla & 3.28 & 1.65 & 3.88 & 2.94 \\
\midrule

\rowcolor{grey!5}
\multicolumn{5}{c}{\textit{Budget: 90\% Tokens}} \\
speedup     & 7.85 & 6.14 & 18.45 & 10.81 \\
interpolate & \underline{3.52} & 1.79 & 4.34  & 3.22  \\
\rowcolor{xbjblue!15}
$\text{AP}_{\text{in}}$       & 3.84 & \underline{\textbf{1.65}} & \underline{\textbf{3.79}}  & \underline{3.09} \\
\midrule

\rowcolor{grey!5}
\multicolumn{5}{c}{\textit{Budget: 80\% Tokens}} \\
speedup     & 8.94 & 6.97 & 21.8  & 12.57 \\
interpolate & 3.86 & 1.84 & 4.25  & 3.32  \\
\rowcolor{xbjblue!15}
$\text{AP}_{\text{in}}$      & \underline{3.44} & \underline{1.75} & \underline{3.92}  & \underline{3.04} \\
\midrule

\rowcolor{grey!5}
\multicolumn{5}{c}{\textit{Budget: 70\% Tokens}} \\
speedup     & 11.92 & 11.28 & 32.98 & 18.73 \\
interpolate & 5.34  & 2.36  & 4.83  & 4.18  \\
\rowcolor{xbjblue!15}
$\text{AP}_{\text{in}}$       & \underline{4.04}  & \underline{2.21}  & \underline{4.42}  & \underline{3.56} \\
\midrule

\rowcolor{grey!5}
\multicolumn{5}{c}{\textit{Budget: 60\% Tokens}} \\
speedup     & 18.04 & 19.85 & 51.34 & 29.74 \\
interpolate & 8.29  & \underline{3.98}  & 6.96  & 6.41  \\
\rowcolor{xbjblue!15}
$\text{AP}_{\text{in}}$       & \underline{5.94}  & 4.38  & \underline{6.78}  & \underline{5.70} \\
\bottomrule
\end{tabularx}
\end{table}

\subsection{Comparative Analysis}
\label{sec:comparative_analysis}
While our previous experiments demonstrate the robustness of deep layers to compression, the fidelity of the initial token selection at the input stage remains the critical bottleneck for information preservation. To rigorously evaluate the effectiveness of our similarity-based method against signal-agnostic approaches, we conduct a controlled experiment under fixed token budgets at the input layer. To ensure a fair comparison, we enforce strict token retention budgets ($K\%$) across all methods, ranging from 90\% down to 60\% of the original sequence length. We compare \affpoo against two established baselines: (1) \textit{Signal-level Speedup}, which accelerates the raw audio via time-stretching prior to encoding; and (2) \textit{Linear Interpolation}, which uniformly downsamples the audio embedding sequence $\mathbf{H}_a^{(0)}$.

As detailed in Table \ref{tab:asr_budget}, signal-agnostic methods suffer rapid degradation as the compression budget tightens. Specifically, \textit{Signal-level Speedup} introduces significant distortion, leading to a drastic increase in WER. While \textit{Linear Interpolation} performs better, it still consistently lags behind our approach. In contrast, \affpoo demonstrates superior robustness. At the aggressive 60\% budget, our method achieves a mean WER of 5.70\%, significantly outperforming other methods. This performance gap highlights a fundamental limitation of fixed-rate compression: speech information is non-uniformly distributed. Rigid downsampling inevitably discards critical phonemic details in dense regions while preserving redundancy in silence. By aggregating tokens based on semantic affinity, our approach aligns better with the information density of the signal, thereby maximizing distinctiveness even under significant compression constraints.

\begin{figure}[t]
    \centering
    \includegraphics[width=1\linewidth]{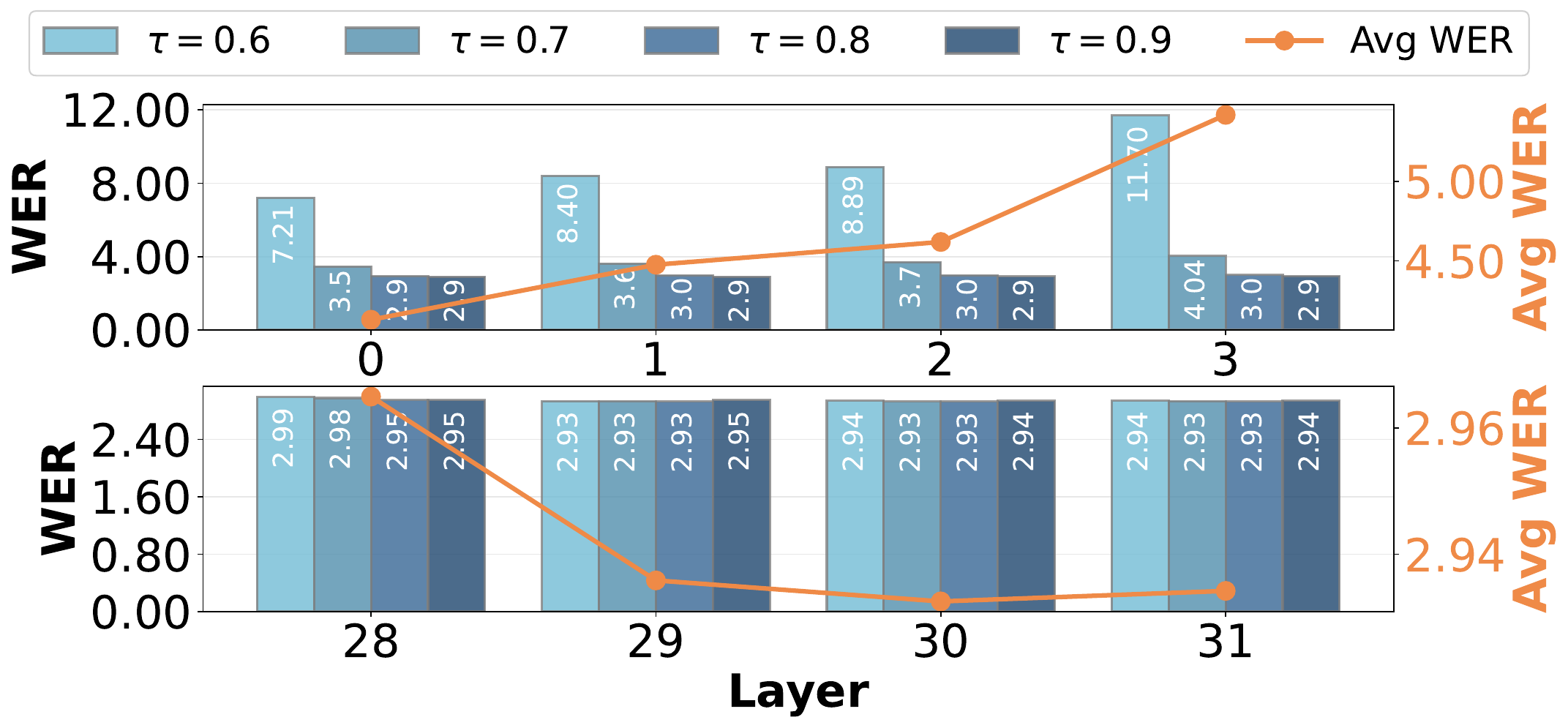}
    \captionsetup{skip=3pt}
    \caption{Layer sensitivity of Qwen2-Audio across early ($l \in [0, 3]$, top) and deep ($l \in [28, 31]$, bottom) layers.}
    \label{fig:qwen_optimal_layer}
\end{figure}

\begin{figure}[t]
    \centering
    \includegraphics[width=1\linewidth]{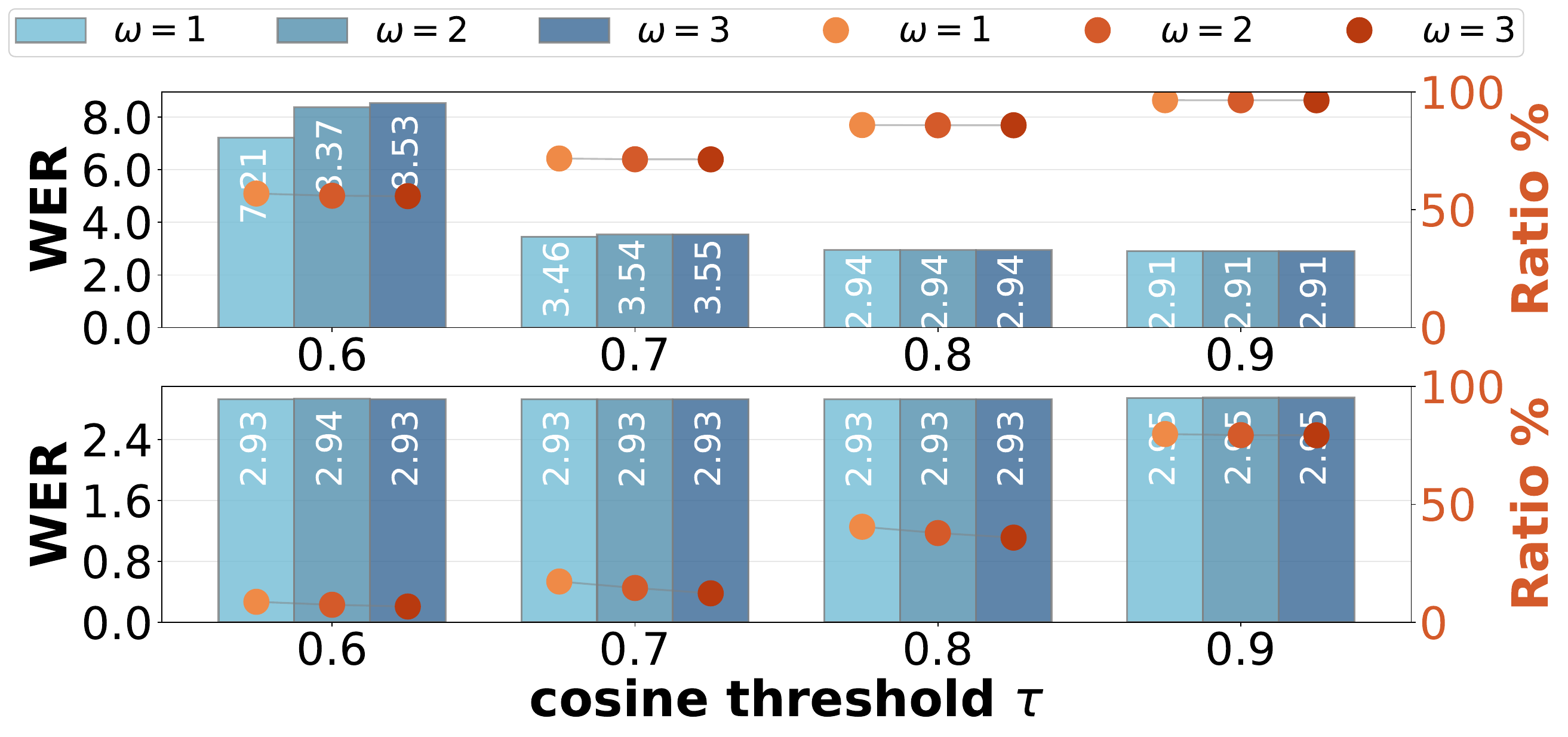}
    \captionsetup{skip=3pt}
    \caption{Lookback window ablation of Qwen2-Audio at input ($l=0$, top) and deep layer ($l=29$, bottom).}
    \label{fig:qwen_optimal_omega}
\end{figure}

\subsection{Parameter Sensitivity}
\label{sec:param_analysis}

We conduct an ablation study to characterize the sensitivity of our method to its three governing hyperparameters: the application layer $l$, the similarity threshold $\tau$, and the lookback window $\omega$. All experiments are performed across three ASR tasks, and we report the average results to ensure generalizability. This analysis identifies the optimal operating points for Qwen2-Audio.

\paragraph{Optimal Layer and Threshold.}
We first examine the impact of the injection point by sweeping across early ($l{\in}[0,3]$) and deep layers ($l{\in}[28,31]$). As illustrated in Figure \ref{fig:qwen_optimal_layer}, the input embedding layer ($l{=}0$) provides the most favorable tradeoff among the shallow layers, effectively reducing sequence length with minimal impact on WER. However, performance degrades notably when compression is applied to the immediate subsequent layers ($l{=}[1,3]$), even at conservative thresholds. In contrast, deep layers exhibit high stability; performance remains robust across a wide range of $\tau$, confirming that the model's final representations can tolerate aggressive compression. These results support an asymmetric configuration: utilizing the input layer for initial reduction and a deep layer for maximizing compression. We observe consistent trends when replicating these experiments on Kimi-Audio (see Appendix \ref{app:kimi_optimal_layer}).

\paragraph{Impact of Lookback Window.}
We further investigate the temporal scope $\omega$. As shown in Figure \ref{fig:qwen_optimal_omega}, the optimal window size depends on the depth. At the input level ($l{=}0$), the method is sensitive to wider windows; increasing $\omega$ beyond 1 leads to higher WER, indicating that strict adjacency is required to preserve acoustic details. Conversely, at deep layers ($l{=}29$), a wider lookback ($\omega{=}3$) improves the compression ratio without compromising accuracy. This justifies the design of DAP, which pairs a local constraint ($\omega{=}1$) at the input with a wider context ($\omega{=}3$) at deeper layers.

\section{Conclusion}

In this work, we explore the potential redundancy of dense tokenization in LSLMs. Through analysis of layer-wise representation evolution, we observe a transition from acoustic details to broader semantic abstractions. Building on these, we introduce \affpoo as a training-free method to reduce computational load and memory usage during inference. Our experiments suggest that leveraging intrinsic feature similarity can be an effective alternative to fixed-rate processing. We hope these findings encourage further investigation into dynamic architectures that more closely align computation with the actual semantic content.

\section*{Limitations}

\paragraph{Scope of Evaluation.} While our experiments demonstrate the efficacy of the proposed compression mechanism on semantics-oriented speech tasks, its impact on fine-grained acoustic details remains underexplored.

\paragraph{Alignment Accuracy.} 
Our oracle analysis relies on forced alignment to find word boundaries. However, these boundaries are approximations and may not perfectly match the actual acoustic transitions in natural speech. This could slightly affect the precision of our compression analysis.

\paragraph{Baseline Comparisons.} 
Since token compression for LSLMs is a new field, there are very few open-source methods available for comparison. Therefore, we compared our approach primarily against standard signal processing techniques.

\bibliography{main}

\appendix

\section{Experimental Details}

\subsection{Detailed Results of Layer-wise Oracle Interventions}
\label{app:oracle_details}

Complementing the redundancy analysis presented in Section \ref{sec:layerwise_analysis}, we provide the comprehensive numerical results corresponding to the visual trends in Figure \ref{fig:chap3_layerwise_main}. Table \ref{tab:qwen2_audio_detailed} and Table \ref{tab:kimi_audio_detailed} detail the performance metrics for Qwen2-Audio and Kimi-Audio, respectively. We report both standard WER and clamped WER (cWER) across all investigated compression operators (Random Drop, Uniform Drop, Uniform Merge) and token retention ratios. These quantitative results substantiate the observed hierarchy of representational density, confirming the distinct stability profiles of deep semantic layers compared to shallow acoustic layers.

\begin{table*}[htbp]
\centering
\small
\setlength{\tabcolsep}{2pt}
\renewcommand{\arraystretch}{1.1}
\caption{\textbf{Detailed WER results for Qwen2-Audio}. Standard WER (WER) and clamped WER (cWER) (in \%) are reported for each operator, token ratio, and layer. The baseline WER is \textbf{1.65}.}
\begin{tabular}{l |c| cccccccccccccc}

\toprule
\multirow{2}{*}{\textbf{Operator}} &
\multirow{2}{*}{\textbf{\makecell{Token\\Ratio}}} &
\multicolumn{2}{c}{\textbf{Layer 0}} &
\multicolumn{2}{c}{\textbf{Layer 5}} &
\multicolumn{2}{c}{\textbf{Layer 10}} &
\multicolumn{2}{c}{\textbf{Layer 15}} &
\multicolumn{2}{c}{\textbf{Layer 20}} &
\multicolumn{2}{c}{\textbf{Layer 25}} &
\multicolumn{2}{c}{\textbf{Layer 30}} \\
\cmidrule(lr){3-4}\cmidrule(lr){5-6}\cmidrule(lr){7-8}\cmidrule(lr){9-10}\cmidrule(lr){11-12}\cmidrule(lr){13-14}\cmidrule(lr){15-16}
& & {\footnotesize WER} & {\footnotesize cWER}
  & {\footnotesize WER} & {\footnotesize cWER}
  & {\footnotesize WER} & {\footnotesize cWER}
  & {\footnotesize WER} & {\footnotesize cWER}
  & {\footnotesize WER} & {\footnotesize cWER}
  & {\footnotesize WER} & {\footnotesize cWER}
  & {\footnotesize WER} & {\footnotesize cWER} \\
\midrule

\multirow{4}{*}{\makecell{Random\\Drop}} & 25.67\% & 177.15 & 65.89 & 421.39 & 76.41 & 382.52 & 70.54 & 454.25 & 67.11 & 154.46 & 26.69 & 2.20 & 1.98 & 1.63 & 1.63 \\
 & 47.57\% & 37.54 & 24.30 & 79.14 & 25.28 & 73.42 & 20.28 & 91.95 & 18.77 & 32.87 & 7.74 & 1.80 & 1.80 & 1.64 & 1.64 \\
 & 76.08\% & 6.33 & 4.88 & 9.41 & 4.84 & 6.87 & 3.71 & 6.92 & 3.26 & 4.97 & 2.55 & 1.69 & 1.69 & 1.65 & 1.65 \\
 & 97.73\% & 1.70 & 1.70 & 1.72 & 1.72 & 2.05 & 1.70 & 1.69 & 1.69 & 1.67 & 1.67 & 1.66 & 1.66 & 1.65 & 1.65 \\
\midrule

\multirow{4}{*}{\makecell{Uniform\\Drop}} & 25.67\% & 141.13 & 58.21 & 274.25 & 61.83 & 244.44 & 53.83 & 300.56 & 49.31 & 102.06 & 18.25 & 2.38 & 1.91 & 1.63 & 1.63 \\
 & 47.57\% & 23.78 & 17.29 & 42.49 & 16.98 & 38.46 & 13.39 & 46.25 & 12.23 & 20.78 & 5.54 & 1.79 & 1.79 & 1.64 & 1.64 \\
 & 76.08\% & 3.23 & 3.23 & 3.41 & 2.84 & 4.00 & 2.56 & 4.28 & 2.30 & 2.43 & 1.97 & 1.70 & 1.70 & 1.65 & 1.65 \\
 & 97.73\% & 1.73 & 1.73 & 1.71 & 1.71 & 1.69 & 1.69 & 1.68 & 1.68 & 1.66 & 1.66 & 1.66 & 1.66 & 1.65 & 1.65 \\
\midrule

\multirow{4}{*}{\makecell{Uniform\\Merge}} & 25.67\% & 183.57 & 58.21 & 169.63 & 56.23 & 72.25 & 33.45 & 90.65 & 28.56 & 17.82 & 5.16 & 1.83 & 1.83 & 1.64 & 1.64 \\
 & 47.57\% & 12.07 & 11.59 & 10.95 & 10.49 & 5.75 & 5.72 & 26.45 & 10.80 & 4.22 & 2.14 & 1.70 & 1.70 & 1.63 & 1.63 \\
 & 76.08\% & 2.50 & 2.50 & 2.39 & 2.39 & 2.01 & 2.01 & 2.19 & 1.99 & 1.65 & 1.65 & 1.65 & 1.65 & 1.63 & 1.63 \\
 & 97.73\% & 1.66 & 1.66 & 1.66 & 1.66 & 1.64 & 1.64 & 1.64 & 1.64 & 1.63 & 1.63 & 1.64 & 1.64 & 1.65 & 1.65 \\
\bottomrule
\end{tabular}

\label{tab:qwen2_audio_detailed}
\end{table*}

\begin{table*}[htbp]
\centering
\small
\setlength{\tabcolsep}{3pt}
\renewcommand{\arraystretch}{1.1}
\caption{\textbf{Detailed WER results for Kimi-Audio}. Standard WER (WER) and clamped WER (cWER) (in \%) are reported for each operator, token ratio, and layer. The baseline WER is \textbf{1.34}.}

\begin{tabular}{l |c| ccccccccccccc}

\toprule
\multirow{2}{*}{\textbf{Operator}} &
\multirow{2}{*}{\textbf{\makecell{Token\\Ratio}}} &
\multicolumn{2}{c}{\textbf{Layer 0}} &
\multicolumn{2}{c}{\textbf{Layer 5}} &
\multicolumn{2}{c}{\textbf{Layer 10}} &
\multicolumn{2}{c}{\textbf{Layer 15}} &
\multicolumn{2}{c}{\textbf{Layer 20}} &
\multicolumn{2}{c}{\textbf{Layer 25}} &
 \\
\cmidrule(lr){3-4}\cmidrule(lr){5-6}\cmidrule(lr){7-8}\cmidrule(lr){9-10}\cmidrule(lr){11-12}\cmidrule(lr){13-14}
& & {\footnotesize WER} & {\footnotesize cWER}
  & {\footnotesize WER} & {\footnotesize cWER}
  & {\footnotesize WER} & {\footnotesize cWER}
  & {\footnotesize WER} & {\footnotesize cWER}
  & {\footnotesize WER} & {\footnotesize cWER}
  & {\footnotesize WER} & {\footnotesize cWER}
 \\
\midrule

\multirow{4}{*}{Random Drop} & 25.67\% & 74.81 & 65.80 & 71.57 & 64.08 & 92.47 & 63.68 & 186.67 & 60.51 & 251.76 & 55.47 & 1.90 & 1.90 \\
 & 47.57\% & 26.09 & 25.53 & 25.39 & 24.14 & 24.64 & 22.75 & 25.80 & 16.50 & 49.52 & 15.76 & 1.59 & 1.59 \\
 & 76.08\% & 4.11 & 4.11 & 3.36 & 3.36 & 3.21 & 3.21 & 2.35 & 2.35 & 3.97 & 2.37 & 1.35 & 1.35 \\
 & 97.73\% & 1.37 & 1.37 & 1.36 & 1.36 & 1.36 & 1.36 & 1.35 & 1.35 & 1.32 & 1.32 & 1.32 & 1.32 \\
\midrule
\multirow{4}{*}{Uniform Drop} & 25.67\% & 66.73 & 61.91 & 64.07 & 59.58 & 74.69 & 57.83 & 130.33 & 52.55 & 199.48 & 48.80 & 2.13 & 1.88 \\
 & 47.57\% & 17.66 & 17.66 & 20.12 & 19.46 & 20.04 & 18.96 & 15.40 & 11.53 & 23.17 & 8.76 & 1.45 & 1.45 \\
 & 76.08\% & 3.11 & 3.11 & 2.59 & 2.59 & 2.68 & 2.68 & 1.98 & 1.98 & 1.69 & 1.69 & 1.36 & 1.36 \\
 & 97.73\% & 1.43 & 1.43 & 1.34 & 1.34 & 1.35 & 1.35 & 1.32 & 1.32 & 1.31 & 1.31 & 1.30 & 1.30 \\
\midrule
\multirow{4}{*}{Uniform Merge} & 25.67\% & 54.32 & 51.48 & 51.52 & 50.40 & 50.33 & 46.91 & 28.78 & 22.81 & 8.09 & 5.73 & 1.47 & 1.47 \\
 & 47.57\% & 13.58 & 13.58 & 15.24 & 15.24 & 14.33 & 13.54 & 5.41 & 5.20 & 1.98 & 1.98 & 1.38 & 1.38 \\
 & 76.08\% & 1.98 & 1.98 & 1.94 & 1.94 & 1.90 & 1.90 & 1.58 & 1.58 & 1.40 & 1.40 & 1.31 & 1.31 \\
 & 97.73\% & 1.34 & 1.34 & 1.35 & 1.35 & 1.33 & 1.33 & 1.35 & 1.35 & 1.31 & 1.31 & 1.32 & 1.32 \\
\bottomrule
\end{tabular}

\label{tab:kimi_audio_detailed}
\end{table*}

\subsection{Detailed Results of Similarity-Driven Interventions}
\label{app:similarity_results}

To supplement the analysis of layer-wise dynamics in Section~\ref{sec4_2_layerwise_dynamics}, we present the full numerical data. Table~\ref{tab:qwen2audio_affpoo_intervention} and Table~\ref{tab:kimiaudio_affpoo_intervention} detail the performance for Qwen2-Audio and Kimi-Audio, respectively. We report WER, clamped WER (cWER), and the resulting audio token retention ratio across varying cosine similarity thresholds ($\tau \in \{0.6, 0.7, 0.8, 0.9\}$) and injection layers. These empirical results corroborate the bimodal stability profile discussed in the main text, highlighting the robustness of input and deep-layer representations to aggressive compression compared to the sensitivity observed in intermediate layers.

\begin{table*}[htbp]
\centering
\small 
\setlength{\tabcolsep}{4.5pt} 

\caption{Performance of Qwen2-Audio across different layers and similarity thresholds $\tau$. Metrics include WER, clamped WER (cWER), and token retention ratio (Ratio). The baseline WER is \textbf{1.65}.}
\label{tab:qwen2audio_affpoo_intervention}

\begin{tabular*}{\textwidth}{@{\extracolsep{\fill}} c|cccccccccccc @{}}
\toprule
\multirow{2}{*}{\textbf{Layer}} & \multicolumn{3}{c}{$\tau = 0.6$} & \multicolumn{3}{c}{$\tau = 0.7$} & \multicolumn{3}{c}{$\tau = 0.8$} & \multicolumn{3}{c}{$\tau = 0.9$} \\
\cmidrule(lr){2-4} \cmidrule(lr){5-7} \cmidrule(lr){8-10} \cmidrule(lr){11-13}
 & \textbf{WER} & \textbf{cWER} & \textbf{Ratio} & \textbf{WER} & \textbf{cWER} & \textbf{Ratio} & \textbf{WER} & \textbf{cWER} & \textbf{Ratio} & \textbf{WER} & \textbf{cWER} & \textbf{Ratio} \\
\midrule
0 & 6.85 & 6.51 & 57.61\% & 1.99 & 1.99 & 74.32\% & 1.63 & 1.63 & 88.60\% & 1.65 & 1.65 & 97.43\% \\
5 & 17.82 & 17.36 & 46.67\% & 3.32 & 3.32 & 66.62\% & 1.66 & 1.66 & 84.71\% & 1.65 & 1.65 & 96.69\% \\
10 & 6.03 & 5.63 & 52.57\% & 2.09 & 2.09 & 70.46\% & 1.66 & 1.66 & 86.74\% & 1.64 & 1.64 & 97.12\% \\
15 & 38.83 & 14.70 & 46.52\% & 3.52 & 2.73 & 67.23\% & 1.65 & 1.65 & 85.92\% & 1.64 & 1.64 & 97.14\% \\
20 & 90.98 & 18.19 & 22.53\% & 6.08 & 2.94 & 45.36\% & 1.68 & 1.68 & 74.12\% & 1.66 & 1.66 & 95.09\% \\
25 & 2.04 & 2.04 & 10.55\% & 1.83 & 1.83 & 26.08\% & 1.68 & 1.68 & 58.06\% & 1.64 & 1.64 & 90.46\% \\
30 & 1.64 & 1.64 & 5.18\% & 1.63 & 1.63 & 10.51\% & 1.65 & 1.65 & 33.89\% & 1.64 & 1.64 & 79.43\% \\
\bottomrule
\end{tabular*}
\end{table*}

\begin{table*}[htbp]
\centering
\small 
\setlength{\tabcolsep}{4.5pt} 

\caption{Performance of Kimi-Audio across different layers and similarity thresholds $\tau$. Metrics include WER, clamped WER (cWER), and token retention ratio (Ratio). The baseline WER is \textbf{1.34}.}
\label{tab:kimiaudio_affpoo_intervention}

\begin{tabular*}{\textwidth}{@{\extracolsep{\fill}} c|cccccccccccc @{}}
\toprule
\multirow{2}{*}{\textbf{Layer}} & \multicolumn{3}{c}{$\tau = 0.6$} & \multicolumn{3}{c}{$\tau = 0.7$} & \multicolumn{3}{c}{$\tau = 0.8$} & \multicolumn{3}{c}{$\tau = 0.9$} \\
\cmidrule(lr){2-4} \cmidrule(lr){5-7} \cmidrule(lr){8-10} \cmidrule(lr){11-13}
 & \textbf{WER} & \textbf{cWER} & \textbf{Ratio} & \textbf{WER} & \textbf{cWER} & \textbf{Ratio} & \textbf{WER} & \textbf{cWER} & \textbf{Ratio} & \textbf{WER} & \textbf{cWER} & \textbf{Ratio} \\
\midrule
0 & 5.39 & 5.39 & 59.27\% & 1.64 & 1.64 & 78.35\% & 1.25 & 1.25 & 95.03\% & 1.29 & 1.29 & 99.51\% \\
5 & 56.51 & 55.91 & 27.21\% & 14.68 & 14.65 & 48.58\% & 1.79 & 1.79 & 73.99\% & 1.32 & 1.32 & 95.73\% \\
10 & 30.00 & 29.44 & 41.27\% & 6.43 & 6.43 & 61.39\% & 1.50 & 1.50 & 82.36\% & 1.30 & 1.30 & 97.80\% \\
15 & 60.93 & 35.30 & 34.16\% & 6.15 & 5.81 & 56.26\% & 1.42 & 1.42 & 81.85\% & 1.31 & 1.31 & 98.37\% \\
20 & 351.99 & 77.06 & 15.15\% & 56.22 & 18.69 & 37.70\% & 2.06 & 1.71 & 68.95\% & 1.30 & 1.30 & 95.48\% \\
25 & 1.67 & 1.67 & 30.35\% & 1.47 & 1.47 & 47.80\% & 1.35 & 1.35 & 70.98\% & 1.28 & 1.28 & 92.10\% \\
\bottomrule
\end{tabular*}
\end{table*}

\section{Extended Analysis on Kimi-Audio}

\subsection{Downstream Task Performance}
\label{app:kimi_main_exp}

\begin{table*}[t]
\centering
\small
\setlength{\tabcolsep}{2.2pt} 
\renewcommand{\arraystretch}{1.18}

\begin{threeparttable}
\caption{\textbf{Main results and ablation studies on Kimi-Audio.} FRR: Final Retention Ratio. Bold indicates improvement over Vanilla, while underlining denotes the best performance within each setting.}
\label{tab:kimi_main}

\begin{tabularx}{\textwidth}{l Y Y | c c c | Y Y Y Y | c c c c |c c Y}
\toprule
\multirow{2}{*}{\textbf{Method}} &
\multicolumn{2}{c|}{\textbf{Scope}} & 
\multicolumn{3}{c|}{Efficiency ($\downarrow$)} &
\multicolumn{4}{c|}{ASR (WER $\downarrow$)} &
\multicolumn{4}{c|}{QA (Acc $\uparrow$)} &
\multicolumn{3}{c}{ST (BLEU $\uparrow$)} \\
\cmidrule(lr){2-3}\cmidrule(lr){4-6}\cmidrule(lr){7-10}\cmidrule(lr){11-14}\cmidrule(lr){15-17}
& $l_{\mathrm{in}}$ & $l_{\mathrm{deep}}$ 
& \makecell{FRR} & \makecell{Pre.\\GFLOPs} & \makecell{FLOPs\\Ratio}
& KES & LSC & LSO & \textit{\textbf{Avg.}}
& OBQA & SDQA & TrQA & \textit{\textbf{Avg.}}
& en2zh & zh2en & \textit{\textbf{Avg.}} \\
\midrule

Vanilla & \nmark & \nmark
& 100.0 & 407.22 & 100.0
& 2.71 & 1.34 & 2.58 & 2.21
& 81.98 & 36.71 & 33.79 & 50.83
& 2.02 & 0.0 & 1.01 \\

\midrule

\rowcolor{xbjblue!15}
\multicolumn{17}{l}{\textit{Setting A: Aggressive ($\tau_{\mathrm{in}}{=}0.80,\tau_{\mathrm{deep}}{=}0.70$)}} \\

$\text{AP}_{\text{in}}$ & \cmark & \nmark
& 86.28 & 351.18 & 86.24
& 2.73 & \textbf{1.30} & \textbf{2.55} & \textbf{2.19}
& \textbf{83.08} & \textbf{37.79} & \textbf{33.79} & \textbf{51.55}
& 2.01 & \textbf{0.02} & \textbf{1.02} \\

$\text{AP}_{\text{deep}}$ & \nmark & \cmark
& 48.81 & 377.39 & 92.68
& 2.83 & 1.48 & 2.68 & 2.33
& \textbf{82.20} & \textbf{37.79} & \textbf{33.79} & \textbf{51.26}
& 2.01 & 0.0 & \textbf{1.01} \\

\rowcolor{grey!5}
DAP & \cmark & \cmark
& \underline{40.71} & \underline{324.64} & \underline{79.72}
& 2.85 & 1.43 & 2.60 & 2.29
& \textbf{82.42} & \textbf{37.61} & 33.69 & \textbf{51.24}
& 1.99 & \textbf{0.02} & \textbf{1.01} \\

\midrule

\rowcolor{xbjblue!15}
\multicolumn{17}{l}{\textit{Setting B: Conservative ($\tau_{\mathrm{in}}{=}0.90,\tau_{\mathrm{deep}}{=}0.80$)}} \\


$\text{AP}_{\text{in}}$ & \cmark & \nmark
& 96.44 & 392.69 & 96.43
& 2.72 & \textbf{1.34} & 2.59 & 2.22
& \textbf{82.64} & \textbf{36.71} & \textbf{33.79} & \textbf{51.05}
& 2.01 & 0.0 & \textbf{1.01} \\

$\text{AP}_{\text{deep}}$ & \nmark & \cmark
& 71.18 & 390.42 & 95.87
& 2.74 & 1.36 & \textbf{2.58} & 2.23
& \textbf{82.64} & \textbf{37.61} & 33.59 & \textbf{51.28}
& \textbf{2.02} & 0.0 & \textbf{1.01} \\

\rowcolor{grey!5}
DAP & \cmark & \cmark
& \underline{68.19} & \underline{376.14} & \underline{92.36}
& 2.77 & 1.36 & 2.59 & 2.24
& \textbf{81.98} & \textbf{36.89} & 33.50 & 50.79
& 2.01 & 0.0 & \textbf{1.01} \\

\bottomrule
\end{tabularx}
\end{threeparttable}
\end{table*}

We benchmark the proposed compression algorithm across ASR, QA, and Speech Translation tasks on Kimi-Audio, with results detailed in Table \ref{tab:kimi_main}. Consistent with the observations on Qwen2-Audio, our method achieves substantial computational savings with negligible impact on semantic preservation.

Specifically, under the \textit{Aggressive} setting, the \textit{Dual Affinity Pooling} (DAP) strategy reduces the Final Retention Ratio (FRR) to $\sim$40.7\%, translating to a significant reduction in prefilling GFLOPs. Despite this compression, the model maintains competitive accuracy on ASR and QA benchmarks compared to the Vanilla baseline.

Note on Speech Translation: We observe near-zero BLEU scores for the zh2en translation task across all settings, including the baseline. This performance stems from the intrinsic limitations of the Kimi-Audio base model in zero-shot cross-lingual generation for this specific direction, rather than artifacts introduced by our compression mechanism. We include these results solely for completeness.

\subsection{Layer Sensitivity Analysis}
\label{app:kimi_optimal_layer}

\begin{figure}
    \centering
    \includegraphics[width=1\linewidth]{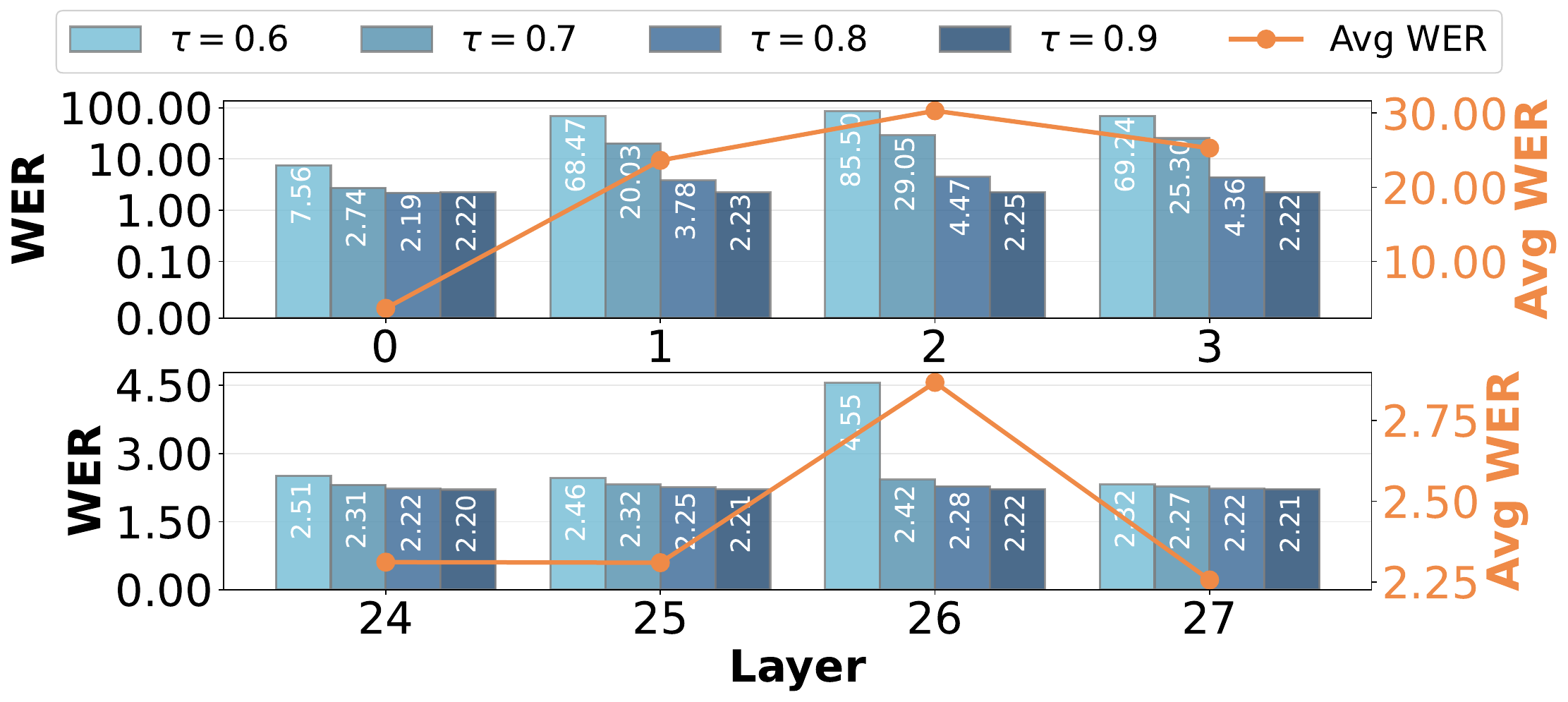}
    \caption{Layer sensitivity of Kimi-Audio across early ($l \in [0, 3]$, top) and deep ($l \in [24, 27]$, bottom) layers. The top plot uses a symlog axis. }
    \label{fig:kimi_layer_sweep}
\end{figure}

We investigate the optimal injection points for compression by conducting a parameter sweep across shallow ($l \in [0, 3]$) and deep ($l \in [24, 27]$) layers on Kimi-Audio. Figure \ref{fig:kimi_layer_sweep} illustrates the WER dynamics under varying similarity thresholds $\tau$.

\paragraph{Shallow Layers ($l \in [0, 3]$).} The input embedding layer ($l=0$) exhibits superior robustness, effectively balancing token reduction with acoustic fidelity. In contrast, injecting compression at immediate subsequent layers ($l \in [1, 3]$) leads to a sharp degradation in performance, confirming that early feature extraction layers are highly sensitive to structural perturbations.

\paragraph{Deep Layers ($l \in [24, 27]$).} Deep representations exhibit remarkable robustness to compression across most operational regimes. Under moderate to conservative thresholds ($\tau \geq 0.7$), all examined layers maintain consistently low WER, corroborating the semantic redundancy hypothesis. However, when applying extremely aggressive compression ($\tau = 0.6$), we observe increased sensitivity at $l=26$, suggesting that while deep layers generally tolerate substantial token reduction, excessively low similarity thresholds may disrupt critical feature alignments even at these abstract representation levels. This finding reinforces the importance of threshold calibration when targeting deep-layer compression.

\section{Benchmark and Evaluation Details}
\label{app:experiment_setups}

\subsection{Benchmark Details}
\label{app:benchmarking}
To comprehensively evaluate the robustness of our proposed compression algorithm, we conduct experiments across three distinct tasks: Automatic Speech Recognition (ASR), Speech Question Answering (QA), and Speech Translation (ST).

\paragraph{Automatic Speech Recognition}
We assess ASR performance using three datasets that vary in language and acoustic complexity. For English, we utilize the Librispeech corpus, reporting results on both the \textit{test-clean} (\textbf{LSC}) and \textit{test-other} (\textbf{LSO}) splits. While \textbf{LSC} (2,620 samples) represents high-quality read speech, \textbf{LSO} (2,939 samples) introduces more challenging acoustic environments. To evaluate multilingual generalization, we employ the test set of KeSpeech (\textbf{KES}), comprising 5,000 Mandarin samples featuring diverse background noises and rich prosodic features. All ASR tasks are evaluated using WER.

\paragraph{Speech Question Answering}
We evaluate semantic understanding using three benchmarks. OpenBookQA (\textbf{OBQA}) consists of 455 long-form audio clips requiring multi-hop reasoning for single-choice questions. We also utilize SDQA-USA (\textbf{SDQA}), which contains 553 real-world spoken queries. Additionally, we incorporate SpeechTriviaQA (\textbf{TrQA}), a dataset sourced from the \textit{TwinkStart} repository. \textbf{TrQA} comprises 1,020 synthetic speech samples, covering open-domain trivia (e.g., general knowledge and pop culture) to test the model's robustness against synthetic prosody. All QA tasks are evaluated using Accuracy (Acc).

\paragraph{Speech Translation}
For translation tasks, we utilize the CoVoST2 corpus, a large-scale multilingual dataset derived from Common Voice. We report results on the English-to-Chinese (\textbf{en2zh}, 15,531 samples) and Chinese-to-English (\textbf{zh2en}, 4,898 samples) directions. Both subsets are characterized by real-world recording conditions, including significant background noise and informal speech. Performance is measured using BLEU scores.

\subsection{QA Evaluation Protocol}
\label{app:evaluation details}

To assess the semantic accuracy of the Speech Question Answering tasks, we employ a model-based evaluation strategy rather than relying solely on exact string matching, which often penalizes valid paraphrases. We deploy \textit{Qwen3-30B-A3B} as our automated evaluator.

The evaluation process involves a comparative analysis where the evaluator is presented with the original question text, the textual response generated by the LSLMs, and the ground-truth reference answer provided by the dataset. The evaluator is instructed to act as an assistant for audio model responses, specifically tasked with determining the correctness of the generated answer by comparing it against the reference. The prompt explicitly directs the model to accept paraphrasing and synonyms while marking the result as correct only if the core information aligns with the reference. The full prompt template is provided below:

\begin{tcolorbox}[
    colback=gray!5!white, 
    colframe=gray!75!black, 
    title=\textbf{Prompt for QA Evaluation Assistant}, 
    boxrule=0.8pt, 
    left=5pt, right=5pt, top=5pt, bottom=5pt, 
    breakable 
]
\ttfamily 
\small 

You are an evaluation assistant for audio model responses.

\textbf{\#\# Task}\\
Determine if an audio model's answer ("audio\_answer") is correct by comparing with reference ("ref"). Accept paraphrasing and synonyms. Mark as correct only if core information matches.

\textbf{\#\# Output only true or false}

\textbf{\#\# Evaluate}\\
Question: \{question\}\\
Audio Answer: \{audio\_answer\}\\
Reference: \{ref\}\
\end{tcolorbox}

\section{Visualizations or Qualitative Results}

\begin{table*}    
\begin{tcolorbox}[colback=gray!2,colframe=gray,title={Decoding Trajectories on Qwen2-Audio under Oracle Interventions}]
\textbf{Ground Truth:} It was strange too that he found an arid pleasure in following up to the end the rigid lines of the doctrines of the church and penetrating into obscure silences only to hear and feel the more deeply his own condemnation.

\vspace{3pt} 
\noindent\hrule
\vspace{3pt} 
\textcolor{xbjred}{\textbf{Random Drop ($R=2$)}}

\textbf{Layer 0:} it was too that he an following up to the end the doctrines of the church and to hear and the own \eos

\textbf{Layer 5:} it was strange too that he an own own own own own own own... \looping

\textbf{Layer 10:} it was strange too that he an and and and and and and and and and and and and and  ... \looping

\textbf{Layer 15:} it was strange too that he he he he he he he he he he he he he he he he he  ... \looping

\textbf{Layer 20:} it was strange too that he he he he he he he he he he he he he he he he he \looping

\textbf{Layer 25 \& Layer 30:} it was strange too that he found an arid pleasure in following up to the end the rigid lines of the doctrines of the church and penetrating into obscure silences only to hear and feel the more deeply his own condemnation \eos

\vspace{3pt} 
\noindent\hrule
\vspace{3pt} 
\textcolor{xbjred}{\textbf{Uniform Drop ($R=2$)}}

\textbf{Layer 0:} it was strange he heard the arduous labours of the church and the obscure toil of the church and the obscure toil of the church and the obscure toil of the church... \looping

\textbf{Layer 5:} it was strange he said he had no desire to hear the end of the story and he was glad to hear that the story was over ... \looping

\textbf{Layer 10:} it was strange he found an arid pleasure in following to the end the obscure of the church and to hear and feel the more own of the narration \eos

\textbf{Layer 15:} it was strange too he he found an arid pleasure in following up to the end the the the the the the the the the the ... \looping

\textbf{Layer 20:} it was strange too that he found an arid pleasure in following up to the end the the the the the the the the the the the \looping

\textbf{Layer 25 \& Layer 30:} it was strange too that he found an arid pleasure in following up to the end the rigid lines of the doctrines of the church and penetrating into obscure silences only to hear and feel the more deeply his own condemnation \eos

\vspace{3pt} 
\noindent\hrule
\vspace{3pt} 
\textcolor{xbjred}{\textbf{Uniform Merge ($R=2$)}}

\textbf{Layer 0:} it was strange that he found an arid in the up to the rigid lines of the of the and to and the more own \eos

\textbf{Layer 5:} it was strange that he found an arid pleasure in following to the end the lines of the book which he had opened at random and which he had read through to the end \eos

\textbf{Layer 10:} \begin{CJK}{UTF8}{gbsn}
它是一个人的心灵的镜子，它是一个人的心灵的镜子，它是一个人的心灵的镜子，它是一个人的心灵的镜子，它是一个人的心灵的镜子 ... \looping
\end{CJK}

\textbf{Layer 15:} it was strange too that he found an arid pleasure in following to the end the rigid lines of the doctrines of the church and penetrating into the arid lines of the doctrines of the church and penetrating into the arid lines of ... \looping

\textbf{Layer 20:} it was strange too that he found an arid pleasure in following to the end the rigid lines of the doctrines of the church and penetrating into obscure silences only to hear and feel the more deeply his own condemnation \eos

\textbf{Layer 25 \& Layer 30:} it was strange too that he found an arid pleasure in following up to the end the rigid lines of the doctrines of the church and penetrating into obscure silences only to hear and feel the more deeply his own condemnation \eos

\end{tcolorbox}
\caption{Decoding Trajectories for an ASR example on Qwen2-Audio under Oracle Interventions. We consolidate Layers 25 and 30 into a single entry, as their decoded outputs are identical, to save space.}
\label{tab:loop_of_qwen}
\end{table*}

\begin{table*}    
\begin{tcolorbox}[colback=gray!2,colframe=gray,title={Decoding Trajectories on Kimi-Audio under Oracle Interventions}]
\textbf{Ground Truth:} A moment before the \textbf{\textcolor{xbjgreen}{ghost}} of the ancient kingdom of the danes had looked forth through the vesture of the hazewrapped city

\vspace{3pt} 
\noindent\hrule
\vspace{3pt} 

\textcolor{xbjred}{\textbf{Random Drop ($R=1$)}}

\textbf{Layer 0:} A ghost of the had for the of the. \eos

\textbf{Layer 5:} Before ghost of the had the wraith of the \eos

\textbf{Layer 10:} Before ghost of the had of the vesture of \eos

\textbf{Layer 15:} The ghost of the ghost of the ghost of the ghost of the ghost of the ghost of the ghost of the ghost of the ghost of  ... \looping

\textbf{Layer 20:} A ghost of the ghost of the ghost of the ghost of the ghost of the ghost of the ghost of the ghost of the ghost of the ghost of  ... \looping

\textbf{Layer 25:} A moment before the ghost of the ancient kingdom of the danes had looked forth through the vesture of the haze wrapped city \eos

\vspace{3pt} 
\noindent\hrule
\vspace{3pt} 
\textcolor{xbjred}{\textbf{Uniform Merge ($R=1$)}}

\textbf{Layer 0:} The the of the city. \eos

\textbf{Layer 5:} The the kingdom of the danes of the city \eos

\textbf{Layer 10:} The ghost of the kingdom had the vesture of the city \eos

\textbf{Layer 15:} The kingdom of the danes had the kingdom of the danes had the kingdom of the danes had the kingdom of the danes had the kingdom of the danes ... \looping

\textbf{Layer 20:} A moment before the \textbf{\textcolor{xbjgreen}{spirit}} of the kingdom of the danes had passed through the veil of the misty night \eos

\textbf{Layer 25:} A moment before the ghost of the ancient kingdom of the danes had looked forth through the vesture of the haze wrapped city \eos

\vspace{3pt} 
\noindent\hrule
\vspace{3pt} 
\textcolor{xbjred}{\textbf{Uniform Drop ($R=1$)}}

\textbf{Layer 0:} A moment later, the man was dead. \eos

\textbf{Layer 5:} A moment the ghost of danes looked through the haze \eos

\textbf{Layer 10:} A moment the ghost of the king of denmark had thrown off the vesture of death. \eos

\textbf{Layer 15:} A moment the ghost of the ancient kingdom of the danes had looked forth through the haze of the city \eos

\textbf{Layer 20:} A moment before the ghost of the ancient kingdom of the danes had looked forth through the vesture of the haze hazed of the danes had looked forth through the vesture of the ... \looping

\textbf{Layer 25:} A moment before the ghost of the ancient kingdom of the danes had looked forth through the vesture of the haze wrapped city \eos

\end{tcolorbox}
\caption{Decoding Trajectories for an ASR example on Kimi-Audio under Oracle Interventions.}
\label{tab:loop_of_kimi}
\end{table*}

\subsection{Decoding Trajectories under Oracle Interventions}
\label{app:oracle_samples}

This section investigates the stability of intermediate representations under different compression operators (Random Drop, Uniform Drop, and Uniform Merge). Tables~\ref{tab:loop_of_qwen} and~\ref{tab:loop_of_kimi} present representative samples from the Librispeech-test-clean dataset showing the decoding trajectories for Qwen2-Audio and Kimi-Audio, respectively.
We observe a vulnerability at intermediate layers where structural perturbations cause divergence, despite the model's ability to recover the ground truth at deeper layers ($l \ge 25$).

\paragraph{Qwen2-Audio (Table~\ref{tab:loop_of_qwen}, we merge Layers 25 and 30 in the table since their decoded outputs are identical for brevity.)}
The model exhibits severe sensitivity to compression in the middle layers ($l \in [5, 20]$). While Random Drop and Uniform Drop primarily trigger repetition loops, Uniform Merge ($R=2$) induces a distinct cross-lingual hallucination at Layer 10, generating fluent but unrelated Chinese text. This suggests that the internal representations in the middle layers are in a highly sensitive transitional state. Unlike deeper layers, these features lack the robustness to withstand structural compression, leading to immediate decoding failures like loops and hallucinations. 

\paragraph{Kimi-Audio (Table~\ref{tab:loop_of_kimi}).} 
Similar instability is observed in Kimi-Audio. Notably, under Uniform Merge ($R=1$), the model displays \textit{semantic drift} at Layer 20 (e.g., paraphrasing "ghost" to "spirit" and "vesture" to "veil"). Unlike the repetition loops caused by Random Drop, this suggests that \textbf{the merged representations retain high-level semantic content} even when exact lexical alignment is temporarily lost. In all cases, the deeper layers demonstrate robustness, effectively correcting these intermediate distortions to reproduce the exact ground truth.

\subsection{Extended Analysis on Semantic Granularity of Aggregated Tokens}
\label{app:more_visualization}

\begin{figure*}[t]
  \centering
  \includegraphics[width=\linewidth]{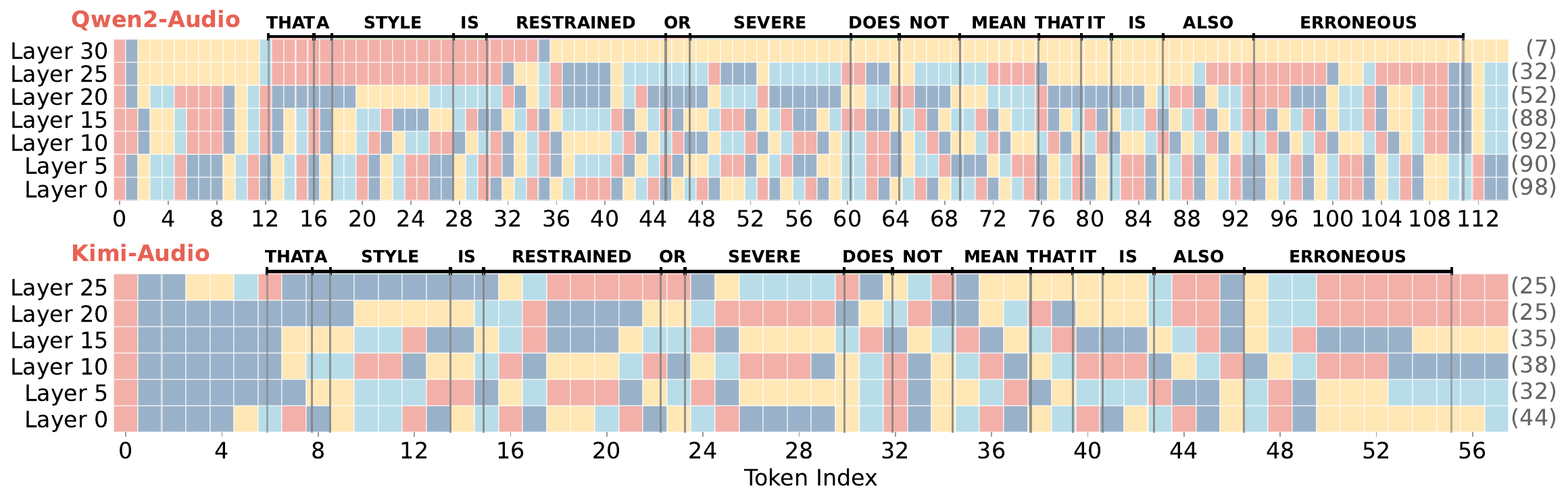}  
  \vspace{1pt} 
  \begin{tcolorbox}[
    colback=gray!5,
    colframe=gray!50,
    title={Decoding Trajectories},
    before=\par\noindent, 
  ]
    \textbf{Ground Truth:} That a style is restrained or severe does not mean that it is also erroneous.

    \vspace{3pt} 
    \noindent\hrule
    \vspace{3pt} 
    
    \textcolor{xbjred}{\textbf{Qwen2-Audio}}
    
    \textbf{Layer 0:} that a style is restrained or severe does not mean that it is also erroneous \eos
    
    \textbf{Layer 5:} that a style is restrained or severe does not mean that it is also erroneous \eos
    
    \textbf{Layer 10:} that a style is restrained or severe does not mean that it is also erroneous \eos
    
    \textbf{Layer 15:} that a style is restrained or severe does not mean that it is also erroneous \eos
    
    \textbf{Layer 20:} that a style is restrained or severe does not mean that it is also erroneous \eos
    
    \textbf{Layer 25:} that a style is restrained or severe does not mean that it is also erroneous \eos

    \vspace{3pt} 
    \noindent\hrule
    \vspace{3pt} 
    
    \textcolor{xbjred}{\textbf{Kimi-Audio}}
    
    \textbf{Layer 0:} That a style is restrained or severe does not mean that it is also erroneous. \eos
    
    \textbf{Layer 5:} A style restrained or severe does not mean that it is also erroneous. \eos
    
    \textbf{Layer 10:} That a style is restrained or severe does not mean that it is also erroneous. \eos
    
    \textbf{Layer 15:} A style is restrained or severe does not mean that it is also erroneous. \eos
    
    \textbf{Layer 20:} That a style is restrained or severe does not mean that it is also erroneous. \eos
    
    \textbf{Layer 25:} That a style is restrained or severe does not mean that it is also erroneous. \eos
      \end{tcolorbox}

  \captionof{figure}{\textbf{Top: Visualization of \affpoo (\(\tau{=}0.7, \omega=3\)) on Qwen2-Audio and Kimi-Audio}. Colors denote merged token groups, and vertical lines mark word boundaries. The right axis indicates the total number of tokens after compression. \textbf{Bottom: ASR transcripts decoded from the compressed representations at the corresponding layers}.}
  \label{fig:cos_example_1}
\end{figure*}


\begin{figure*}[t]
  \centering
  \includegraphics[width=\linewidth]{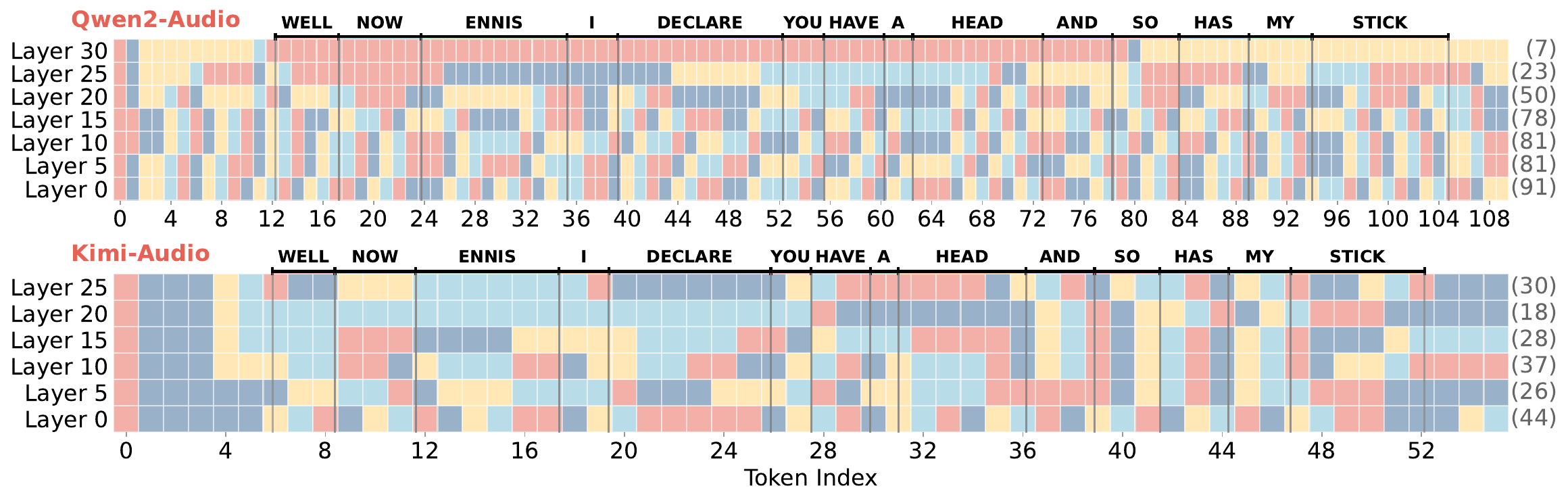}  
  \vspace{1pt} 
  \begin{tcolorbox}[
    colback=gray!5,
    colframe=gray!50,
    title={Decoding Trajectories},
    before=\par\noindent, 
  ]
    \textbf{Ground Truth:} Well now ennis i declare you have a head and so has my stick.

    \vspace{3pt} 
    \noindent\hrule
    \vspace{3pt} 
    
    \textcolor{xbjred}{\textbf{Qwen2-Audio}}
    
    \textbf{Layer 0:} well now ennis i declare you have a head and so has my stick \eos
    
    \textbf{Layer 5:} well now ennis i declare you have a head and so has my stick \eos
    
    \textbf{Layer 10:} well now ennis i declare you have a head and so has my stick \eos
    
    \textbf{Layer 15:} well now ennis i declare you have a head and so has my stick \eos
    
    \textbf{Layer 20:} well now ennis i declare you have a head and so has my stick \eos
    
    \textbf{Layer 25:} well now ennis i declare you have a head and so has my stick \eos

    \vspace{3pt} 
    \noindent\hrule
    \vspace{3pt} 
    
    \textcolor{xbjred}{\textbf{Kimi-Audio}}
    
    \textbf{Layer 0:} Well now, Innes, I declare you have a head, and so has my stick. \eos
    
    \textbf{Layer 5:} Well now, I declare you have a head, so has my stick. \eos
    
    \textbf{Layer 10:} Well now, Innes, I declare you have a head, and so has my stick. \eos
    
    \textbf{Layer 15:} Well now, Innes, you have a head, and so has my stick. \eos
    
    \textbf{Layer 20:} \textbackslash"Ah, now, Nairn, I'll have you to remember that I'm a gentleman.\textbackslash" \textbackslash"I'll remember, I'll remember, and so has my stick. \eos
    
    \textbf{Layer 25:} Well now, Innes, I declare you have a head, and so has my stick. \eos
      \end{tcolorbox}

  \captionof{figure}{\textbf{Top: Visualization of \affpoo (\(\tau{=}0.7, \omega=3\)) on Qwen2-Audio and Kimi-Audio}. Colors denote merged token groups, and vertical lines mark word boundaries. The right axis indicates the total number of tokens after compression. \textbf{Bottom: ASR transcripts decoded from the compressed representations at the corresponding layers}.}
  \label{fig:cos_example_2}
\end{figure*}


\begin{figure*}[t]
  \centering
  \includegraphics[width=\linewidth]{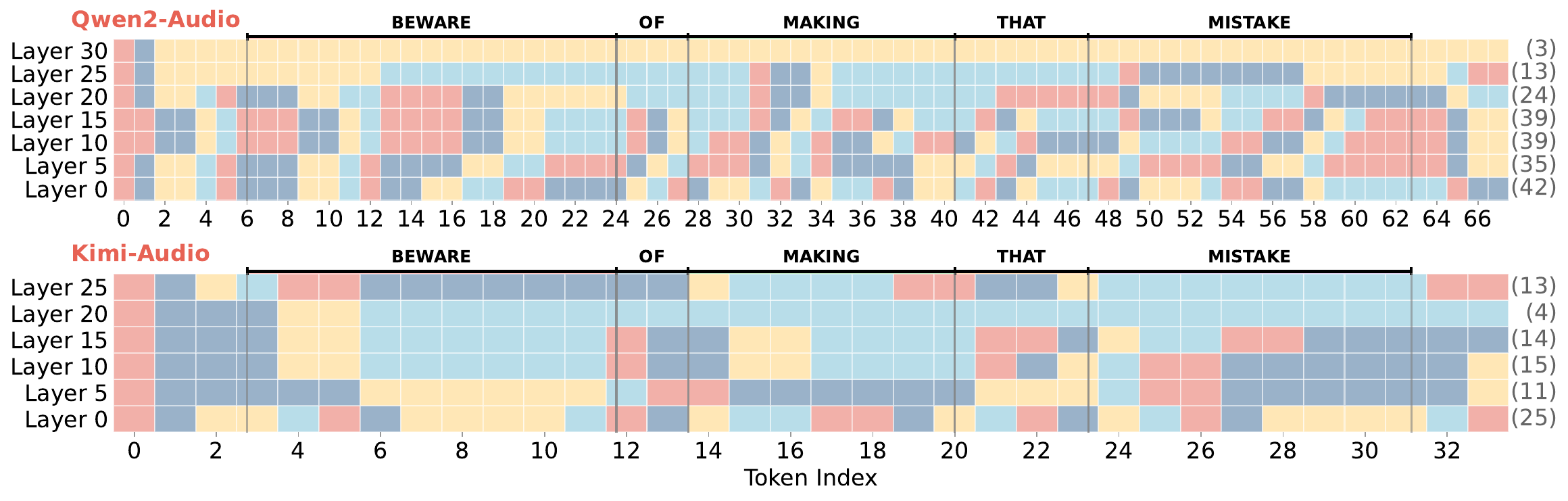}  
  \vspace{1pt} 
  \begin{tcolorbox}[
    colback=gray!5,
    colframe=gray!50,
    title={Decoding Trajectories},
    before=\par\noindent, 
  ]
    \textbf{Ground Truth:} Beware of making that mistake.

    \vspace{3pt} 
    \noindent\hrule
    \vspace{3pt} 
    
    \textcolor{xbjred}{\textbf{Qwen2-Audio}}
    
    \textbf{Layer 0:} beware of making that mistake \eos
    
    \textbf{Layer 5:} beware of making that mistake \eos
    
    \textbf{Layer 10:} beware of making that mistake \eos
    
    \textbf{Layer 15:} beware of making that mistake \eos
    
    \textbf{Layer 20:} beware of making that mistake \eos
    
    \textbf{Layer 25:} beware of making that mistake \eos

    \vspace{3pt} 
    \noindent\hrule
    \vspace{3pt} 
    
    \textcolor{xbjred}{\textbf{Kimi-Audio}}
    
    \textbf{Layer 0:} Beware of making that mistake. \eos
    
    \textbf{Layer 5:} Beware of that mistake. \eos
    
    \textbf{Layer 10:} Beware of making that mistake. \eos
    
    \textbf{Layer 15:} \textbackslash"Be ware of making that mistake. \eos
    
    \textbf{Layer 20:} \textbackslash"Be ware of be ware of be ware of be ware of be ware of be ware of be ware of be ware of be ware of be ware of be ware of be ware of... \looping
    
    \textbf{Layer 25:} \textbackslash"Beware of making that mistake. \eos
    \end{tcolorbox}

  \captionof{figure}{\textbf{Top: Visualization of \affpoo (\(\tau{=}0.7, \omega=3\)) on Qwen2-Audio and Kimi-Audio}. Colors denote merged token groups, and vertical lines mark word boundaries. The right axis indicates the total number of tokens after compression. \textbf{Bottom: ASR transcripts decoded from the compressed representations at the corresponding layers}.}
  \label{fig:cos_example_3}
\end{figure*}

To further substantiate the observations in Section \ref{sec:qualitative_analysis} regarding the layer-wise evolution of representation density, we provide additional visualizations of \affpoo applied to Qwen2-Audio and Kimi-Audio. Figures \ref{fig:cos_example_1}, \ref{fig:cos_example_2}, and \ref{fig:cos_example_3} illustrate the token aggregation patterns ($\tau=0.7, \omega=3$) alongside the corresponding decoded transcripts across varying depths.

Consistent with our main findings, the visualizations reveal a distinct hierarchy in token granularity. In shallow layers ($l \le 5$), the aggregation groups are fragmented and short. As depth increases, these groups expand significantly. By the deep layers ($l \ge 25$), single tokens frequently span entire phrases or multi-word clauses. For instance, in Figure \ref{fig:cos_example_1}, Qwen2-Audio compresses the input of 115 audio tokens into fewer than 10 tokens at Layer 30 while maintaining a perfect transcript.

The decoding trajectories further highlight the robustness of our similarity-based compression at the input and deep layers, while exposing the sensitivity of intermediate representations. Both models consistently recover the ground truth at the input layer ($l=0$) and deep layers ($l \ge 25$). This validates our asymmetric design choice in \textit{Dual Affinity Pooling} (DAP). However, we observe transient instability in the middle layers ($l \in [15, 20]$), particularly in Kimi-Audio. As seen in Figure \ref{fig:cos_example_3}, compression at Layer 20 triggers a repetition loop (\textit{be ware of...}''), and in Figure \ref{fig:cos_example_2}, it induces a hallucination (\textit{Ah, now, Nairn...}''). These errors vanish at Layer 25, suggesting that while intermediate layers are structurally fragile to aggregation, the final semantic layers reorganize these features into a highly robust, compressible format. These qualitative results reinforce our quantitative findings: redundancy in LSLMs is not uniform but structurally organized, transitioning from acoustic redundancy in shallow layers to semantic redundancy in deep layers.

\end{document}